\documentclass[acmtog,authorversion]{acmart}
\usepackage{booktabs} 
\pdfoutput=1

\setcitestyle{square}

\usepackage[ruled]{algorithm2e} 

\SetAlFnt{\small}
\SetAlCapFnt{\small}
\SetAlCapNameFnt{\small}
\SetAlCapHSkip{0pt}
\IncMargin{-\parindent}

\acmJournal{TOG}

\acmYear{2018}
\acmVolume{37}
\acmNumber{6}
\acmMonth{11}
\acmArticle{245}
\acmDOI{10.1145/3272127.3275013}

\setcopyright{acmcopyright}

\usepackage{graphicx}
\usepackage{rotating}
\usepackage{array}
\usepackage{float}
\usepackage{diagbox}
\usepackage{multirow}
\usepackage{subcaption}
\usepackage{tikz}
\usepackage{subcaption}
\usepackage{tikz}
\usepackage{pgfplots}
\usepackage{adjustbox}
\usepackage{animate}
\usetikzlibrary{spy,calc}

\makeatletter
\@namedef{ver@everyshi.sty}{}
\makeatother
\definecolor{myblue}{rgb}{.1,0.3,0.9}

\usepackage{colortbl}
\usepackage{tabularx}
\usepackage{booktabs}
\usepackage{mathtools}
\usepackage{amsmath}
\usepackage{amssymb}

\definecolor{myorchid}{RGB}{150,10,80}
\definecolor{darkgreen}{RGB}{10,150,10}
\definecolor{darkred}{RGB}{100,150,20}
\definecolor{YellowOrange}{RGB}{255,127,39}

\newcommand{\final}[1] {{#1}}
\newcommand{\ignore}[1]{}

\usepackage{booktabs}
\usepackage{xfrac}
\usepackage{tablefootnote}
\usepackage{colortbl}
\usepackage{tabularx}
\definecolor{rowblue}{RGB}{220,230,240}

\usepackage{gensymb}
\usepackage{hyperref}
\usepackage{bm}

\begin{document}
\title{DeepLens: Shallow Depth Of Field From A Single Image}

\author{Lijun Wang}
\authornote{Work done while interining at Adobe Research.}
\affiliation{
    \institution{Dalian University of Technology}
}
\email{wlj@mail.dlut.edu.cn}
\author{Xiaohui Shen}
\affiliation{
    \institution{ByteDance AI Lab}
}
\email{shenxiaohui@bytedance.com}
\author{Jianming Zhang}
\email{jianmzha@adobe.com}
\author{Oliver Wang}
\email{owang@adobe.com}
\author{Zhe Lin}
\email{zlin@adobe.com}
\affiliation{
    \institution{Adobe Research}
}
\author{Chih-Yao Hsieh}
\email{hsieh@adobe.com}

\author{Sarah Kong}
\email{sakong@adobe.com}
\affiliation{
    \institution{Adobe Systems}
}

\author{Huchuan Lu}
\affiliation{
    \institution{Dalian University of Technology}
}
\email{lhchuan@dlut.edu.cn}

\begin{abstract}
We aim to generate high resolution shallow depth-of-field (DoF) images from a single all-in-focus image with controllable focal distance and aperture size.
To achieve this, we propose a novel neural network model comprised of a depth prediction module, a lens blur module, and a guided upsampling module.
All modules are differentiable and are learned from data.
To train our depth prediction module, we collect a dataset of 2462 RGB-D images captured by mobile phones with a dual-lens camera, and use existing segmentation datasets to improve border prediction.
We further leverage a synthetic dataset with known depth to supervise the lens blur and guided upsampling modules.
The effectiveness of our system and training strategies are verified in the experiments.
Our method can generate high-quality shallow DoF images at high resolution, and produces significantly fewer artifacts than the baselines and existing solutions for single image shallow DoF synthesis.
Compared with the iPhone portrait mode, which is a state-of-the-art shallow DoF solution based on a dual-lens depth camera, our method generates comparable results, while allowing for greater flexibility to choose focal points and aperture size, and is not limited to one capture setup.




\end{abstract}

%
%


%
%


\begin{teaserfigure}\centering
\animategraphics[autoplay,loop,height=3.0cm]{1}{figures/anims/1_}{0}{2} 
\animategraphics[autoplay,loop,height=3.0cm]{1}{figures/anims/2_}{0}{2}
\animategraphics[autoplay,loop,height=3.0cm]{1}{figures/anims/3_}{0}{1}
\animategraphics[autoplay,loop,height=3.0cm]{1}{figures/anims/4_}{0}{2}
\animategraphics[autoplay,loop,height=3.0cm]{1}{figures/anims/5_}{0}{1}
    \caption{\label{fig:teaser}Each high resolution, shallow DoF image shown here was synthetically generated from an all-in-focus input image with a single forward pass through a neural network system. Users can freely interact with the system by modifying the focal plane and aperture size. Please view zoomed-in on screen, and use Acrobat Reader to watch the refocusing animations. \textit{Photo credits: Franco Vannini, Jim Liestman, wombatarama, and European Space Agency.}}
\end{teaserfigure}

\maketitle

\section{Introduction}
The shallow depth-of-field (DoF) effect is an important technique in photography which draws the viewers' attention to the region of focus by blurring out the rest of the image. 
Typically, these images are captured with expensive, single-lens reflex (SLR) cameras and large-aperture lenses. 
This makes their acquisition less accessible to casual photographers.
In addition, once captured, these images can not be easily edited to refocus onto other regions, or to change the amount of defocus. 

In this work, we propose to train a neural network system which generates synthetic shallow DoF effects on all-in-focus photos captured by ordinary cameras or mobile devices. 
%
%
%
Our network is composed of a depth prediction module for single image depth estimation, a lens blur module for predicting spatially-varying blur kernels, and a guided upsampling module for generating high-resolution shallow DoF images.

\final{The whole network is fully differentiable and therefore allows end-to-end training. However, the required training data (all-in-focus and shallow DoF image pairs) are hard to capture for diverse scenes with non-static content. 
One potential solution for data collection is to generate ground truth shallow DoF images of real scenes with existing rendering methods that leverage depth maps. 
However, as image-based methods suffer from various artifacts (\emph{c.f.} Fig.~\ref{fig:lensblur}), the rendered ground truth contains artifacts.
An alternative solution is to adopt artifact-free synthetic data, but then the challenge is to generalize this data to real world images.
Based on these discussions, we propose to train the network in a piece-wise manner. Specifically, since depth prediction requires semantic understanding of the scene, we train the depth prediction module on real images to allow for generalization.
In comparison, rendering DoF effects based on the predicted depth mainly involves low-level operations which are consistent across real and synthetic data. 
To alleviate artifacts, the lens blur and guided upsampling modules are therefore jointly trained on high-quality synthetic images.}   


To train our depth prediction module, we collect a new dataset consisting of 2462 RGB-D images of diverse content from real-world environments. 
The RGB-D images are acquired by the dual-lens camera on an iPhone. 
Compared with existing RGB-D datasets \cite{Silberman2012nyuv2,Geiger2012CVPR,SrinivasanWSRN17}, our proposed dataset contains more diverse contents and is more focused on everyday scenes and common photography subjects.
We further leverage existing salient object datasets~\cite{cheng2015global,wang2017learning} to augment the depth training with an auxiliary foreground segmentation task, which improves the quality at object boundaries, and makes the depth prediction module better generalize to diverse content and image types.

The lens blur module together with the guided upsampling module generates high resolution shallow DoF images based on predicted depth maps.
To jointly train the two modules, we introduce a synthetic dataset consisting of scenes with known depth and occlusion information, which can be used to render shallow DoF supervision without artifacts.
We further randomly distort the known depth during training to make our lens blur module and guided upsampling module robust to inaccurate depth maps. 
After training, the guided upsampling module can be recurrently applied on the output of the lens blur module to generate high-resolution results of arbitrary resolutions (we show results up to 2k in this paper). 
We conduct user studies to show that our results significantly outperform existing single image-based solutions, and are comparable to DoF images rendered by the iPhone portrait mode, which requires depth inputs from a dual-lens camera. 
Our method can be used with any camera, and allows users to control the focal point and blur amount at interactive rates (\emph{e.g.}, 0.7s for 2048 resolution).
We further show that our method can be applied in more general scenarios such as historic photos and even paintings.

In summary, we present the following contributions:
\begin{enumerate}
\item A learning-based DoF rendering method that can be trained to 
produce fewer artifacts than existing hand-designed lens blur methods, while being more robust to inaccurate depth maps. 
This is trained with a loss designed to concentrate the training on depth discontinuity boundaries. 

\item A memory-efficient network architecture that performs the lens blur operation in a compressed feature space and recurrently applies guided upsampling to generate results at high resolution.

\item A carefully-designed training scheme using a combination of real and synthetic data for the general single image shallow DoF synthesis. 
\end{enumerate}

\section{Related Work}
\label{sec:related}
\begin{figure*}[t]
    \centering
    \includegraphics[width=1\linewidth]{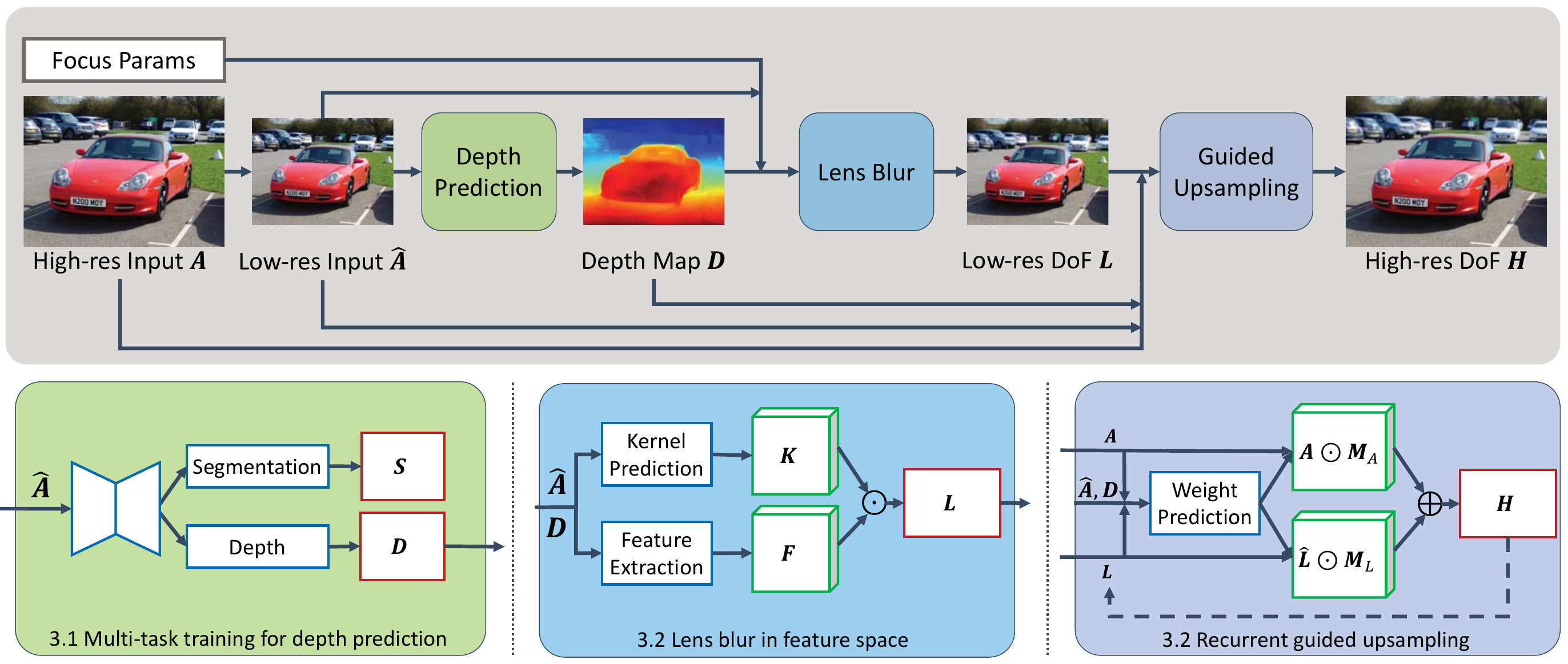}
    \caption{\textbf{Method Overview}. At a high level (top), we predict a low resolution depth map $\bm{D}$ from the all-in-focus input $\bm{\hat{A}}$. $\bm{D}$, $\bm{\hat{A}}$, and the focus parameters are used to generate a low resolution shallow DoF image $\bm{L}$, which is then upsampled into a final high resolution output $\bm{H}$. In more detail (bottom), the depth prediction module takes $\bm{\hat{A}}$ and jointly predicts depth $\bm{D}$ and segmentation $\bm{S}$ maps, both of which have supervision. The lens blur module computes spatially varying kernels $\bm{K}$ which are applied to a reduced dimension feature map $\bm{F}$ to produce $\bm{L}$. The guided upsampling module recurrently super-resolves the DoF image by a factor of $\times 2$ until it reaches the original resolution. Red blocks indicate supervision, blue blocks are learnable modules, and green blocks are intermediate feature maps. See the supplementary materials for more details of the network architecture. \textit{Photo credits: grassrootsgroundswell.}}
    \label{fig:arch}
    \vspace{-2mm}
\end{figure*}

Depth reconstruction is a classic problem in computer vision with a large body of literature. 
Here, we review learning-based single image depth estimation methods, as well as related works that address depth in the context of lens blur effects. 
We will also cover related works in lens blur rendering and defocus estimation.

\paragraph{Single Image Depth Estimation.}
CNNs have proven to be powerful tools to learn strong scene priors for single image depth prediction, usually by training to reconstruct the output of depth cameras~\cite{eigen2014depth,liu2015deep,laina2016deeper}. 
However, depth cameras often have limitations in terms of range, image quality, portability, and resolution.
As a result, models trained on raw depth data often only work in limited scenarios (\textit{e.g.}, indoor rooms~\cite{Silberman2012nyuv2}, street scenes~\cite{Geiger2012CVPR}, landmarks~\cite{li2018megadepth}). 
In ~\cite{pratul2018}, aperture stacks are used as supervision for depth prediction, which demonstrates its effectiveness on flower images and indoor scenes. However, only using aperture supervision can suffer from depth ambiguities, especially with diverse scenes, and we found in our experiments that we were not able to generate high quality results by using aperture supervision alone . 

\paragraph{Multi-Images Depth Estimation.}
Using correspondences across multiple images is the most common way to acquire depth information, and several methods have used this in the context of shallow DoF effects. 
For example, depth can be created from focal stacks on mobile phones~\cite{suwajanakorn2015depth}, stereo camera pairs~\cite{barron2015fast}, small baseline burst image capture~\cite{joshi2014micro,ha2016high}, or multiple frames of video~\cite{klose2015sampling}.
These methods all perform a two stage process, where the pre-computed depth and an in-focus image is used to synthesize shallow DoF results. 
As opposed to this, we compute depth from a single image, and introduce a differential shallow DoF renderer, which is trained to be robust to errors in estimated depth. 

\paragraph{Lens Blur Effects.} 
Lens blur effects that operate on a single RGB-D image cannot generate an exact result, as defocus blur requires seeing \emph{behind} objects. Existing methods that approximate the lens blur can be generally classified into object space and image space. 
Object space methods \cite{haeberli1990accumulation,lee2010real} explore ray tracing and real camera models, and are more effective in rendering realistic shallow DoF images, but most of these methods are time-consuming and entail 3D scene representation which is hard to achieve. 
Image space methods \cite{lee2009real,kraus2007depth,yang2016virtual} operate on a single image, and cast shallow DoF effects rendering as a post-processing step with gathering and spreading operations. 
These methods are more efficient and can be converted to a fixed but differentiable layer, as has been done in the compositional aperture rendering~\cite{pratul2018}. 
However, this approach uses fixed convolutional kernels (discs) at discrete depths and blends them to generate a final image. 
This can lead to results that suffer from intensity leakage at depth discontinuities, and cannot handle foreground occluders in the defocused regions well.
As opposed to these methods, our approach uses a lens blur module that is motivated by real lens optics, and can produce effects such as an out-of-focus foreground, which is more challenging to generate.

Current high end mobile phones, such as the iPhone and Google Pixel 2, have a ``portrait mode'' application, which simulates shallow DoF images.
These are based off of depth measurements from multiple views or dual-pixels, or from a simpler foreground segmentation mask. Moreover, users are unable to control the focal point or the blur amount in the current portrait mode application.
In contrast, our method allows users to modify the shallow DoF parameters interactively, and can generate results with comparable quality \emph{without} a dedicated hardware depth module.

\section{Method}

While it is possible to cast the shallow DoF rendering problem as a black-box input-to-output regression problem~\cite{Isola2017ImagetoImageTW}, we found this approach did not generate satisfactory results.
Instead, we propose an approach motivated by physical models that consists of three modules: \textbf{depth prediction}, \textbf{lens blur}, and \textbf{guided upsampling}.
Fig.~\ref{fig:arch} shows an overview of our approach. 
We now describe each step in the model, but please refer to the supplementary materials for detailed network architectures.

\subsection{Depth Prediction}
\label{sec:depthmethod}
We use a fully convolutional network architecture for depth prediction, trained on our newly collected RGB-D dataset (see Section~\ref{sec:iphonedata}). 
To further increase the generalization ability of our model across different image types, we augment the training with an additional foreground segmentation task, using \cite{cheng2015global,wang2017learning}
The auxiliary segmentation task also enforces more accurate object boundary delineation in depth prediction, which is essential to shallow DoF rendering.

The depth prediction module is shown in Fig.~\ref{fig:arch}. 
It consists of an encoder and a decoder followed by multi-task heads.
The encoder extracts features using the first 14 residual blocks (from Conv1 to Res4f) of the ResNet-50~\cite{he2016deep} network pretrained on ImageNet~\cite{deng2009imagenet}, with atrous convolution layers on the last block to preserve feature resolution.
It is then followed by a four-level pyramid pooling layer~\cite{zhao2017} with bin sizes of $1\times 1$, $2\times 2$, $4 \times 4$ and $8\times 8$, respectively. 
The decoder then upsamples the feature maps with three $\times 2$ bilinear upsampling layers, with skip connections from the encoder at each level.
At the end of the decoder, the network is split into two small multi-task heads, each consisting of three convolutional layers. 
These heads generate depth prediction and foreground segmentation respectively. 



Shallow DoF rendering only relies on depth ordinal information, making the reconstruction of the \emph{absolute} depth values unnecessary. \final{We therefore invert the depth to compute the disparity which is further normalized to the range of $[0,1]$ by the minimum and maximum values. Both the ground truth and the network prediction are normalized in the same manner.
We use "depth prediction" to refer to the generation of these normalized inverted depth maps in the rest of the text.}

The network is then trained in a multi-task manner using the following loss function:
\begin{equation}\label{eq:depth_loss}
    J_d(\theta_{d}) = \|\bm{D} - \bm{D}^g\|_1 + \gamma \times \|\bm{S} - \bm{S}^g\|_1,
\end{equation}
where $\bm{D}$ and $\bm{S}$ denote the predicted depth map and foreground segmentation map, respectively; $\bm{D}^g$ and $\bm{S}^g$ are the corresponding ground truth; $\gamma$ represents a trade-off parameter to balance the two tasks. 
We note that $\bm{D}^g$ and $\bm{S}^g$ come from different datasets, and each mini-batch during training contains images from both datasets. During the test phase, we only maintain the depth prediction stream, while the foreground segmentation stream is discarded.

\input{src/fig/lensblur.tex}

\subsection{Lens Blur Rendering}
Given the predicted depth map along with the original image, existing image space or object space rendering methods could be utilized to approximate lens blur effects~\cite{lee2010real, pratul2018, yang2016virtual}.
However, these methods often produce artifacts around object boundaries.
Therefore, to obtain higher quality results, we use a differentiable neural network to approximate the spatially varying lens blur kernel, and train this network on a new artifact-free synthetic dataset (see Section~\ref{sec:syn}).
We compare the results of our lens blur approach with the ones generated by other lens blur rendering methods in Fig.~\ref{fig:lensblur}, where all methods are given the same input image and depth map. We can see that our learned lens blur module produces more accurate results with much fewer artifacts.

One challenge with this approach is that it is prohibitive to directly predict and apply the learned spatially varying lens blur kernel on a large image due to memory consumption. 
Consider an input image of $H\times W$ resolution and a maximum blur kernel size of $k\times k$. This would require computing an $H\times W \times k^2$ kernel tensor. 
Applying a spatially varying filter with a kernel size $k=65$ on a 1280$\times$1280 3-channel image  would require $1280^2 \times 65^2 \times 4 = 25.79$ GB memory, and 19.34 GFLOPS at inference time\footnote{Filtering the image with spatially varying kernels amounts to an element-wise multiplication between the kernel and the image tensor. Since both tensors are of size $1280\times 1280 \times 65^2 \times 3$, the total computational complexity is $1280\times 1280 \times 65^2 \times 3 = 19.34$ GFLOPS.} .
To reduce memory and computation, we operate on a lower resolution image ($h \times w$ where  $h<H$ and $w<W$) and rely on the subsequent recurrent guided upsampling module to generate high-resolution results.
Furthermore, the lens blur module operates as $1\times 1$ filters in a \emph{learned feature space} with $c << k^2$ dimensions, instead of directly applying blur in the image space.
As a consequence, we only have to predict a kernel tensor of $h \times w \times c$.


\subsubsection{Lens Blur in Learned Feature Space.}
The lens blur module is comprised of a feature extraction network and a kernel prediction network. 
The feature extraction network takes as input the low resolution all-in-focus RGB image of size $h\times w$ and produces a feature map of size $h\times w \times c$.
It consists of four convolution layers interleaved by ReLU units. 
To capture multi-scale features, the final feature maps are obtained by concatenating all intermediate feature maps, which mainly encodes low-level features around each local image patch (\emph{e.g.}, color, blur, texture, \emph{etc}).

Given the depth map, focal depth and aperture radius, the kernel prediction network is used to infer $1 \times 1$ blur kernels for each image location by generating a $h\times w \times c$ kernel tensor, which is then applied to the extracted feature map of each color channel to render a shallow DoF effect. Formally, given the extracted feature $\bm{F}_i \in \mathbb{R}^{h \times w \times c}$ for color channel $i \in \{R,G,B\}$, and the predicted kernel tensor
$\bm{K} \in \mathbb{R}^{h \times w \times c}$, the shallow DoF image is rendered by:
\begin{equation}\label{eq:blur}
    \bm{L}_i(x,y) = \sum_{j=1}^c \bm{K}(x,y,j) \times \bm{F}_i(x,y,j),
\end{equation}
where $\bm{L}_i(x,y)$ denotes the color value at location $(x,y)$ of the rendered DoF image. \final{Since the kernel prediction and feature extraction networks are jointly trained with shallow DoF supervision, they learn in a data-driven fashion to generate desired kernels and feature space, leading to high quality shallow DoF results.}

As the predicted depth itself may not be accurate, relying only on the depth to infer blur kernels increases the risk of producing unpleasant artifacts. 
Therefore, the final kernel prediction network also takes the all-in-focus image as input, which allows it to correct for errors in the predicted depth map based on the image context. This results in more robust inference. 
Please refer to the supplementary material for more network architecture details.

\subsubsection{Control Parameters for Focal Depth and Aperture Size.}
The lens blur network requires both the focal depth and aperture radius to generate an output image.
We encode focal depth with a signed depth map $\bm{\tilde{D}}$ by subtracting the focal depth from the predicted normalized depth map: $\bm{\tilde{D}}=\bm{D} - d_f$, where $d_f$ denotes the focal depth (0 implies that depth is in focus).

In order to encode the aperture size into the network input, one could simply tile the desired aperture size and concatenate it with the signed depth map.
However, we found that this approach led to unsatisfactory results, as it was difficult for the network to learn the correlation between the blur amount and the value of aperture radius. 
We instead propose a simpler and more effective alternative moviated by the physics of DoF, where the network takes the signed depth map as input, and is trained to produce shallow DoF images for the \textit{largest} aperture radius handled by our method, $r^m$.
To successfully do this, the network must learn to add different amounts of blur based on the distance to the focal plane.
At test time, we then allow the user to control the blur amount by \emph{scaling} the input depth map with the ratio $\alpha=\frac{r^*}{r^m}$, where $r^*\leq r^m$ denotes the target aperture radius. 
We find that this method leads to much more robust training and allows for continuous aperture radius control at test time.



\subsubsection{Shallow DoF Upsampling.}
One could employ an out-of-the-box super-resolution method to increase output resolution, but in our case we have additional information in the original high resolution all-in-focus image.
Most existing super-resolution methods do not assume this additional information and therefore cannot easily restore the clear in-focus regions.
Another possible solution would be to use a joint guided upsampling approach~\cite{he2010guided} with the high-resolution all-in-focus image as guidance.
However, as this method is not specifically designed for our task, it results in a sub-optimal solution, oversmoothing the in-focus region, and creating artifacts in the defocused region where the guidance image has strong texture (Fig.~\ref{fig:upsample}).

A key observation is that we should upsample the in-focus regions by leveraging the all-in-focus guidance image, while the blurry defocused regions should instead rely more on the low-resolution shallow DoF image. 
Motivated by this, we propose a guided upsampling network, which extracts features separately from the low-resolution space (\emph{i.e.}, the low-resolution input image, the predicted DoF image and the depth map) and the high-resolution space (\emph{i.e.} the high-resolution all-in-focus image). 
The low-resolution features are upsampled to the same resolution of the high-resolution ones using bilinear upsampling. 
The network then predicts two high-resolution spatial weight maps $\bm{M}_A$ and $\bm{M}_L$, from the concatenation of high and low-resolution features.
The final high-resolution shallow DoF image $\bm{H}$ is achieved through a weighted combination of the high-resolution all-in-focus image and the upsampled shallow DoF image:

\begin{equation}\label{eq:upsample}
\bm{H} = \bm{M}_A \odot \bm{A} + \bm{M}_L \odot \bm{\hat{L}},
\end{equation}
where $\odot$ denotes element-wise multiplication, $\bm{A}$ denotes the high-resolution all-in-focus image, and $\bm{\hat{L}}$ represents the shallow DoF image upsampled from the low-resolution version $\bm{L}$~(Equation~\eqref{eq:blur}) using bilinear interpolation.
As shown in Fig.~\ref{fig:upsample}, our guided upsampling module can render more pleasant results than single image super resolution and guided filtering.

\input{src/fig/upsample.tex}

\subsubsection{Joint Training.}
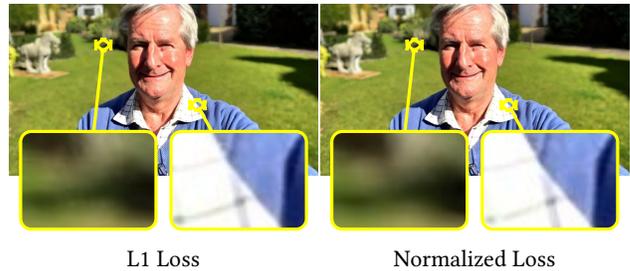
\begin{figure}
	\centering
	\definecolor{mycyan}{RGB}{255,255,0}
	\newcommand\myspy[4]{\spy [mycyan,
		spy connection path={\draw[very thick, mycyan] (tikzspyonnode) -- (tikzspyinnode);},
		every spy on node/.append style={very thick},
		every spy in node/.append style={very thick}] on ({#1},{#2}) in node [left] at ({#3},{#4})}
	\newcommand{\myincludegraphics}[1]{\includegraphics[width=0.48\linewidth,clip,trim=0 0 0 200]{#1}}
	\tabcolsep0.5pt \renewcommand{\arraystretch}{0.5}
	\begin{tabular}{ccc}
				
		\begin{tikzpicture}[spy using outlines={magnification=9,width=1.8cm, height=1.3cm, connect spies, rounded corners},inner xsep=0cm]	
		\node (O) [] {
			\myincludegraphics{{{figures/loss/l1/0}}}
		};
		\myspy{-0.8}{.6}{-.1}{-1.2};
		\myspy{0.45}{-.2}{1.9}{-1.2};		
		\end{tikzpicture} &
		
		\begin{tikzpicture}[spy using outlines={magnification=9,width=1.8cm, height=1.3cm, connect spies, rounded corners},inner xsep=0cm]		
		\node (O) [] {
			\myincludegraphics{{{figures/loss/ours/0}}}
		};
		\myspy{-0.8}{.6}{-.1}{-1.2};
		\myspy{0.45}{-.2}{1.9}{-1.2};		
		\end{tikzpicture} \\
		\\
		L1 Loss &Normalized Loss
	\end{tabular}
    \caption{Comparison of different loss functions. Our normalized loss reduces the artifact on the focus/defocus boundary, as well as in high contrast regions in the background. \textit{Photo credit: Herry Lawford}}
	\label{fig:lossfunctions}
\end{figure}

Since both lens blur and guided upsampling modules are fully differentiable, we jointly train the two parts by minimizing the differences between the produced low and high-resolution results and the corresponding ground truth. 
However, for rendering the shallow DoF images, the most challenging parts of the scene are at object boundaries with depth discontinuities. These regions only constitute a small portion of the whole image but significantly contribute to the perceived quality. 
Traditional $\ell_1$ and $\ell_2$ losses are unable to capture these challenging regions, leading to severe artifacts around object boundaries, as shown in Fig.~\ref{fig:lossfunctions}. 
\final{
Therefore, we propose to use the following normalized loss:
\begin{equation}\label{eq:normalized_fl}
    J_{f}(\bm{X}|\theta) =  \frac{\sum_i w_i L(X_i,Y_i)}{\sum_i w_i},
\end{equation}
where $\bm{X}$ and $\bm{Y}$ denote prediction and ground truth, respectively; $i$ indicates spatial index; $L$ represents some pixel-wise loss; and $\theta$ denotes network parameters to be optimized. The loss weight
$w_i$ for location $i$ is computed as the local error $(|X_i - Y_i|)^{\alpha}$. The hyper-parameter $\alpha$ is empirically set to $1.5$ and works well. 
In our experiments, we adopt the absolute difference as the pixel-wise loss $L$. By substituting $L$ and $w_i$, Equation~\eqref{eq:normalized_fl} can be written as \begin{equation}\label{eq:normalized_fl2}
    J_{f}(\bm{X}|\theta) =  \frac{\sum_i |X_i-Y_i|^{\alpha+1}}{\sum_i |X_i-Y_i|^{\alpha}}.
\end{equation}
Unlike a standard $\ell_p$ loss, this normalization places more emphasis on pixels with large errors, \emph{i.e.}, hard examples, and ignores easy cases 
(Fig.~\ref{fig:lossfunctions}). 
}

\ignore{
Therefore, we consider the following focal loss proposed in \cite{Lin2017FocalLF}:
\begin{equation}\label{eq:fl}
    FL(\bm{X}|\theta) =  \frac{1}{N}\sum_{i=1}^{N} w_i L(X_i,Y_i),
\end{equation}
where $\bm{X}$ and $\bm{Y}$ denote prediction and ground truth, respectively; $i$ indicates spatial index; $L$ represents some pixel-wise loss; $N$ denotes the number of pixels; and $\theta$ denotes network parameters to be optimized. The loss weight
$w_i$ for location $i$ is computed according to the local error $(s|X_i - Y_i|)^{\alpha}$, with $s>0$ and $\alpha>1$ being hyper-parameters.

However, due to the large normalization factor (\emph{e.g.}, $N= 1280^2$ for 1280 resolution), the loss has almost entirely diminished when the network has only addressed the easy cases, still leaving severe problems at the boundaries in the final result. 
To alleviate this issue, we propose the following normalized focal loss which is better suited to pixel-level training:

The major difference from the original focal loss is that it normalizes the loss value using the sum of weights $w_i$. 
This allows the network to focus on the hard examples and not be affected by the imbalanced ratio between hard and easy examples (Fig.~\ref{fig:lossfunctions}). In our experiments, we adopt the absolute differences as the pixel-wise loss $L$. Equation~\eqref{eq:normalized_fl} can then be rewritten as 
\begin{equation}\label{eq:normalized_fl2}
    J_{f}(\bm{X}|\theta) =  \frac{\sum_i |X_i-Y_i|^{\alpha+1}}{\sum_i |X_i-Y_i|^{\alpha}}.
\end{equation}
}

Both the lens blur and guided upsampling modules are then jointly trained using the normalized loss as follows:
\begin{equation}
\arg  \min_{\theta_l, \theta_h} J_f(\bm{L}|\theta_l) + \beta \times J_f(\bm{H}|\theta_h),
\end{equation}
where the low-resolution prediction $\bm{L}$ and its high-resolution counterpart $\bm{H}$ are computed using Equation \eqref{eq:blur} and \eqref{eq:upsample}, $\theta_l$ and $\theta_h$ denote network parameters of lens blur and guided upsampling module, respectively. 
The loss is optimized by stochastic gradient solver. More implementation details about network training and architecture can be found in the supplementary material.

\section{Datasets}
In order to train the proposed method, we require training samples consisting of an all-in-focus and shallow DoF image pair, the corresponding focal depth, aperture size, and ground truth depth map.
As collecting data with all of these characteristics is challenging, we combine different sources of data to supervise the different components of our method.

\subsection{iPhone Depth Dataset}
\label{sec:iphonedata}
\begin{figure*}
    \centering
    \footnotesize
    \tabcolsep0.5pt \renewcommand{\arraystretch}{0.5}
    \begin{tabular}{cccccc}
    \includegraphics[width=0.16\textwidth]{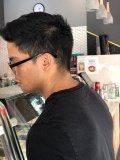}     
    &\includegraphics[width=0.16\textwidth]{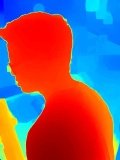}
    &\includegraphics[width=0.16\textwidth]{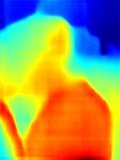}
    &\includegraphics[width=0.16\textwidth]{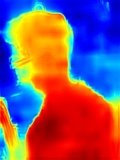}
    &\includegraphics[width=0.16\textwidth]{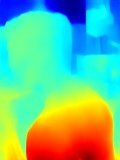}
    &\includegraphics[width=0.16\textwidth]{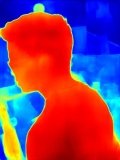}
    \\
    \includegraphics[width=0.16\textwidth]{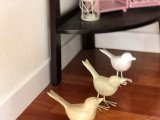}     
    &\includegraphics[width=0.16\textwidth]{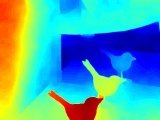}
    &\includegraphics[width=0.16\textwidth]{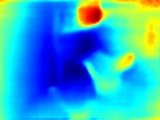}
    &\includegraphics[width=0.16\textwidth]{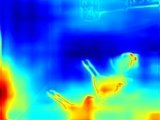}
    &\includegraphics[width=0.16\textwidth]{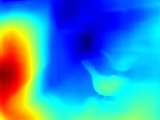}
    &\includegraphics[width=0.16\textwidth]{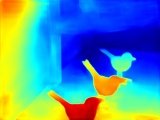}
    \\
    Image &GT &\cite{laina2016deeper} &Aperture &MegaDepth~\shortcite{li2018megadepth} &Ours
    \end{tabular}
    \caption{Predicted depth on iPhone test set.}
    \label{fig:depth_test}
\end{figure*}

Existing depth prediction and shallow DoF datasets are very limited in terms of scene diversity (\textit{e.g.}, indoor rooms~\cite{Silberman2012nyuv2}, street scenes~\cite{Geiger2012CVPR}, and flower images~\cite{SrinivasanWSRN17}), which are not suitable for DoF effect synthesis in general scenes. 
To facilitate network training, we collect a new data set of RGB-D images using the dual-lens camera that comes with recent iPhones. 
We developed a custom iOS app that saves the images and depth maps from the iPhone camera, and deployed it to 15 users who helped capture a wide variety of scenes in different cities. 
After manually filtering out inaccurate depth samples, we finally obtain 2462 RGB-D images, (depth maps are computed at a lower 768$\times$1024 resolution).
The dataset is randomly split into a training set with 2262 images and a test set with 200 images. 
A manual check is performed to guarantee that there is no near-duplicate samples between the training and test sets. Fig.~\ref{fig:depth_test} shows two examples of the test images and depth maps.

Given the input RGB images and depth maps, we could render approximate shallow DoF images for supervision using a ray-tracing based rendering method~\cite{yang2016virtual}. 
However the rendered DoF images are not artifact-free, due to i) the noise in the captured depth maps, and ii) the depth discontinuity artifacts caused by the missing appearance information of the occluded background that is needed for the exact rendering.
Consequently, we found that training our rendering modules on this dataset led to sub-optimal results. Thus, we resort to a synthetic dataset for training our lens blur module and guided upsampling module.


\subsection{Synthetic Shallow DoF Image Data Set}\label{sec:syn}
In order to achieve artifact-free training data, we introduce a cost-effective synthetic shallow DoF dataset consisting of both all-in-focus images and depth maps where we can ensure that i) the depth maps are accurate  ii) the appearance information of occluded regions is available, and iii) the focal plane is known.

To this end, we select 300 training images from our iPhone depth data set, which do not contain any foreground objects and serve as the background images. 
For foreground objects, we augment the matting data set from \cite{xu2017deep} with additional manually annotated data, which in total consists of 3662 foreground images and the corresponding binary masks. 
To synthesize an all-in-focus image, we randomly select one background image and two foreground objects, and composite them together, forming three depth planes. 
The background depth are either all zero or smoothly vary from $b_0$ to $b_1$ either horizontally or from bottom to top, where $b_0>b_1 \in [0,1]$. 
The depth within two foreground planes are homogeneous with values $f_0>f_1$, where $f_0$ and $f_1$ denote the depth values of objects within the front and middle planes, respectively, and are randomly selected from the range of $[0,1]$,
Given the synthesized image, depth map, and a random focal depth, we then render shallow DoF effects using the method~\cite{yang2016virtual} \emph{separately} on the three image planes and composite them together from back to front to generate the final shallow DoF image. \final{Since the depth values within each foreground region are assumed to be homogeneous, depth discontinuities only occur at object boundaries.}
This compositing based approach effectively avoids discontinuity artifacts that arise from image-based shallow DoF rendering approximations, as content \emph{behind} the foreground occluders is known, and can be correctly modeled by ray tracing rather than hallucinated.

\begin{figure}
	\centering
	\small
	\resizebox{.99\linewidth}{!}{
	\tabcolsep0.3pt \renewcommand{\arraystretch}{0.5}
	\begin{tabular}{cccc}
		\includegraphics[width=0.23\textwidth,clip,trim=0 30 0 0]{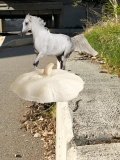}  &
		\includegraphics[width=0.23\textwidth,clip,trim=0 30 0 0]{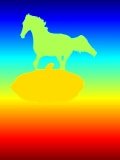} &
		\includegraphics[width=0.23\textwidth,clip,trim=0 30 0 0]{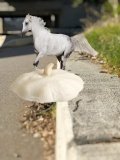}&
		\includegraphics[width=0.23\textwidth,clip,trim=0 30 0 0]{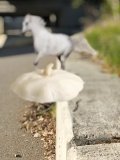} 
		\\
		\includegraphics[width=0.23\textwidth,clip,trim=0 30 0 0]{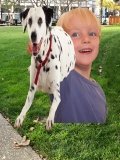}  &
		\includegraphics[width=0.23\textwidth,clip,trim=0 30 0 0]{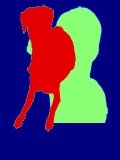} &
		\includegraphics[width=0.23\textwidth,clip,trim=0 30 0 0]{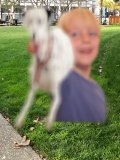}&
		\includegraphics[width=0.23\textwidth,clip,trim=0 30 0 0]{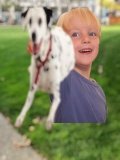} 
		\\
		Input & Depth & DoF & DoF
	\end{tabular}
	}
	\caption{Examples from the synthetic dataset.}
	\label{fig:syn}
\end{figure}

\begin{figure*}
    \centering
    \footnotesize
    \tabcolsep0.5pt \renewcommand{\arraystretch}{0.5}
    \begin{tabular}{ccccc}
    \includegraphics[width=0.19\textwidth]{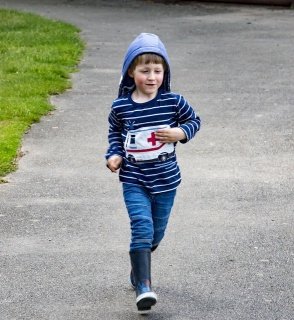}     
    &\includegraphics[width=0.19\textwidth]{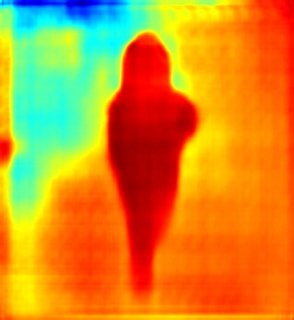}
    &\includegraphics[width=0.19\textwidth]{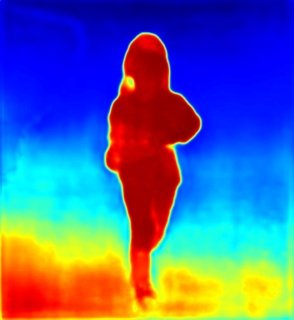}
    &\includegraphics[width=0.19\textwidth]{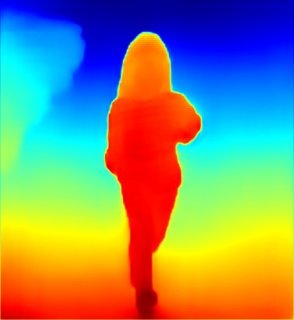}
    &\includegraphics[width=0.19\textwidth]{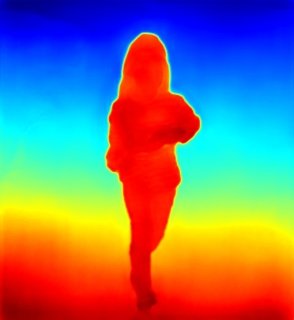}
    \\
    \includegraphics[width=0.19\textwidth]{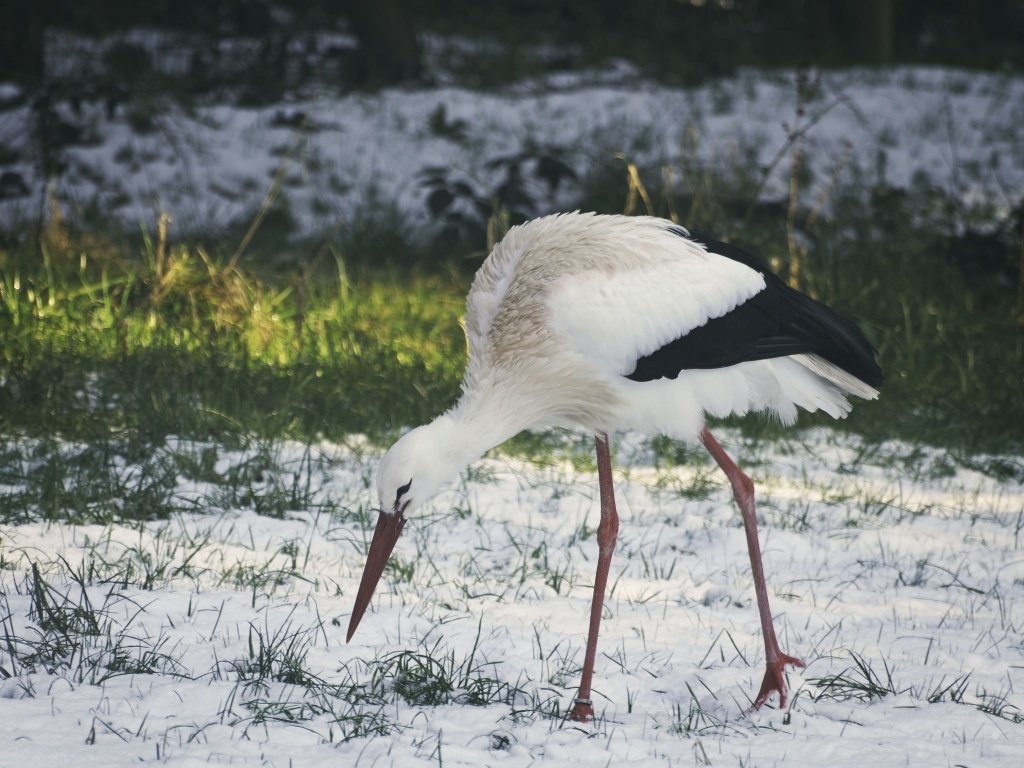}     
    &\includegraphics[width=0.19\textwidth]{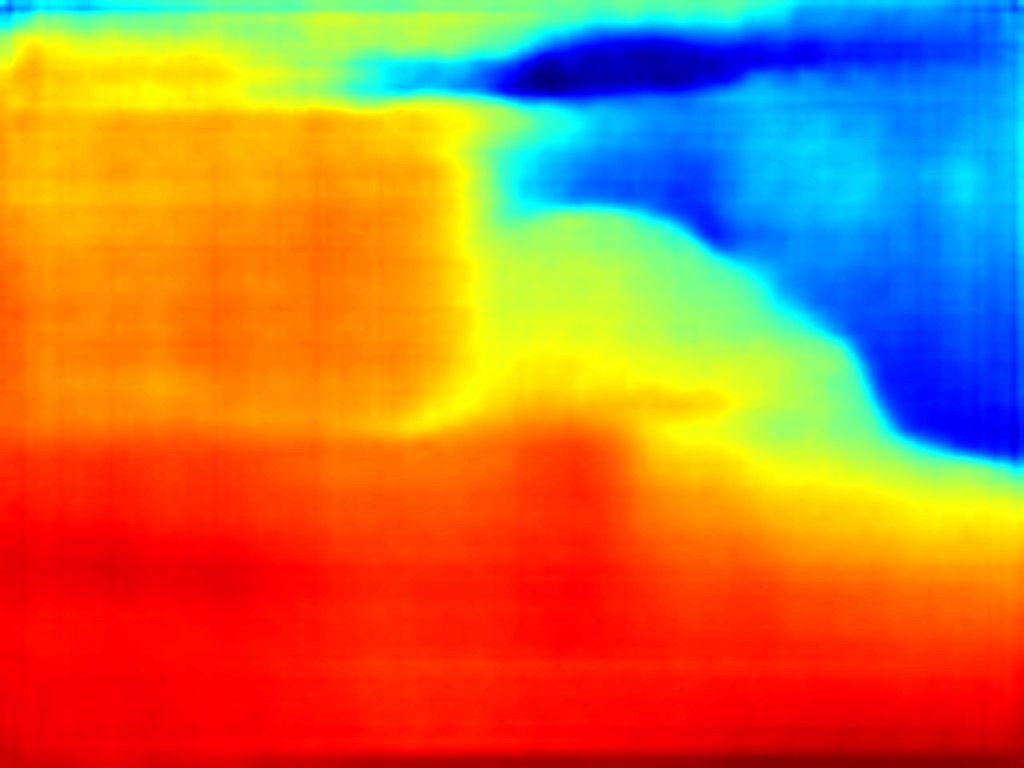}
    &\includegraphics[width=0.19\textwidth]{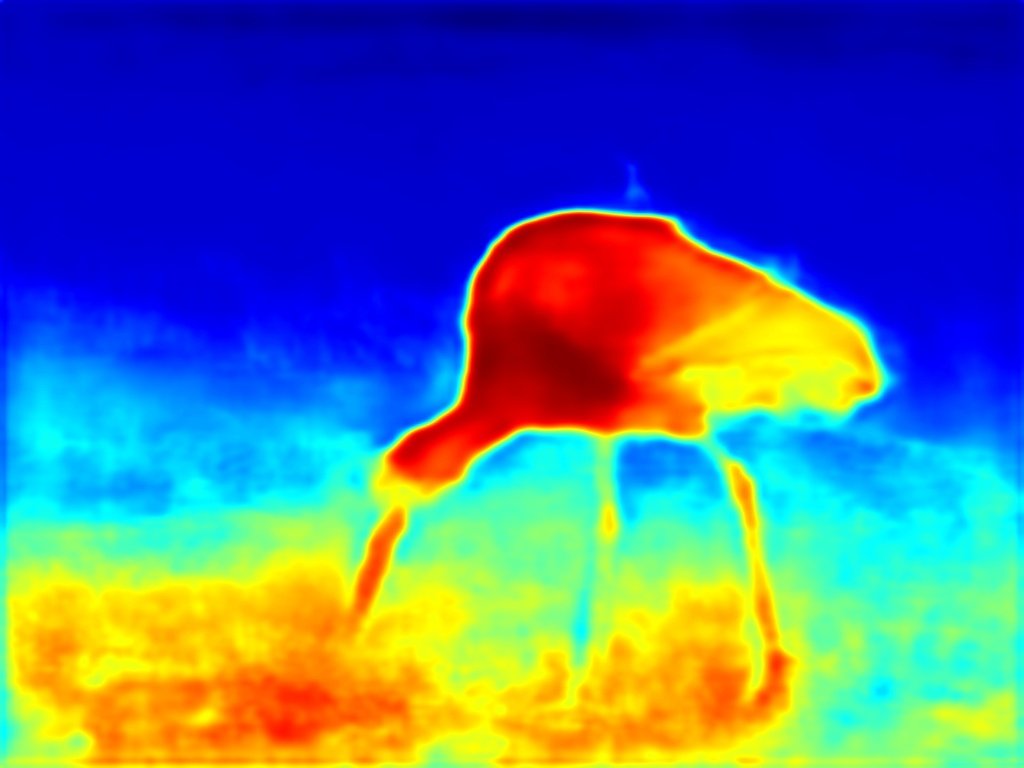}
    &\includegraphics[width=0.19\textwidth]{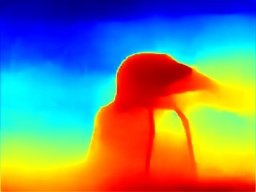}
    &\includegraphics[width=0.19\textwidth]{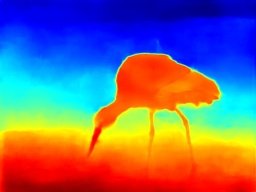}
    \\
    Image &\cite{laina2016deeper} &Aperture &MegaDepth~\shortcite{li2018megadepth} &Ours
    \end{tabular}
    \caption{Predicted depth of images from internet. \textit{Photo credits: Ashley Buttle and Torsten Behrens}}
    \label{fig:depth_web}
\end{figure*}

According to the above rules, we have synthesized 18K training samples and 500 test ones. 
Fig.~\ref{fig:syn} shows two examples of the synthetic data. 
While this data does not look realistic, we find that the lens blur and guided upsampling module trained on the synthetic data can generalize well to real images.
This is because rendering shallow DoF effects mainly involves \emph{low-level} image operations, so the high level scene context is less important. 
We also find randomly corrupting the foreground depth plane through dilation or erosion to be an effective data-augmentation strategy during training.
As shown in our experiments, training on corrupted depth can further improve the robustness against inaccurate depth prediction and forces the lens blur network to make prediction also based on input images rather than depth only.

\final{Besides using corrupted depth, we have also explored training with depth maps generated by our depth prediction module. However, since our depth prediction module is trained on real images, it fails to generalize well to synthetic scenes. 
As a consequence, we train an additional depth prediction module on the synthetic data set.  
We then use the synthetic predicted depth as input to train the lens blur and guided upsampling. 
Our experiments (Section~\ref{sec:ablative}) show that training with predicted depth achieves similar performance to training with corrupted depth.}

\section{Experiments}

\subsection{Depth Prediction}\label{sec:depth}

\begin{figure*}[t]
	\centering
	\small
	\tabcolsep0.3pt \renewcommand{\arraystretch}{0.5}
	\begin{tabular}{cccc}
		\includegraphics[height=0.15\textwidth,clip,trim=50 400 00 125]{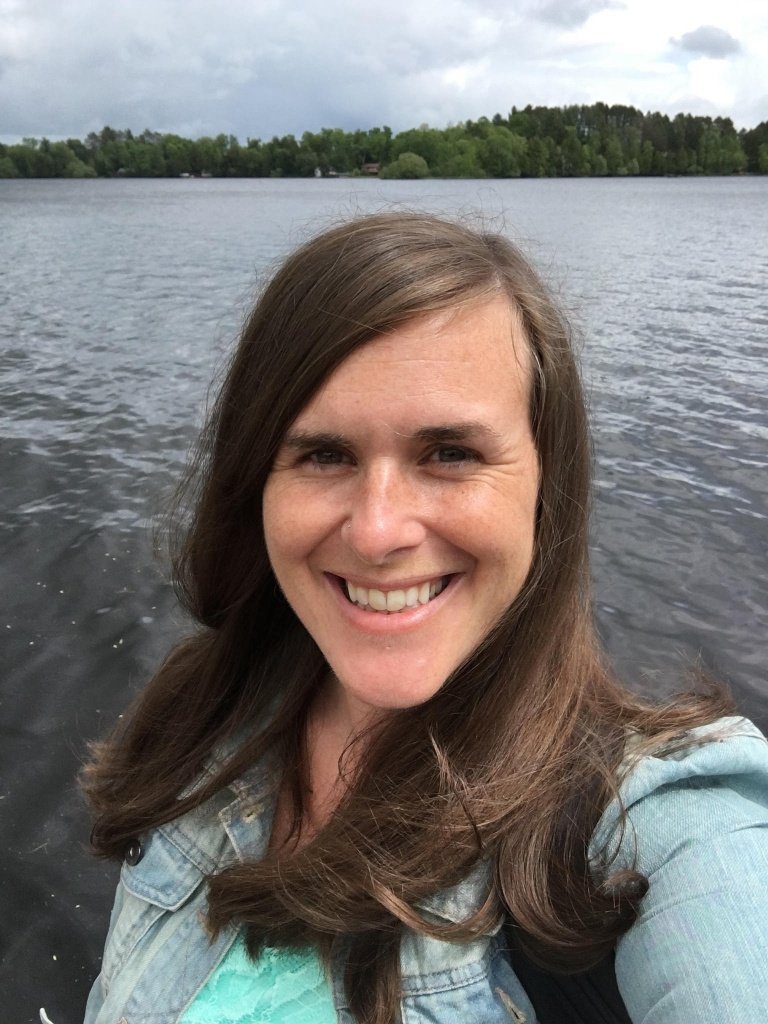}    
        &\includegraphics[height=0.15\textwidth,clip,trim=50 150 100 500]{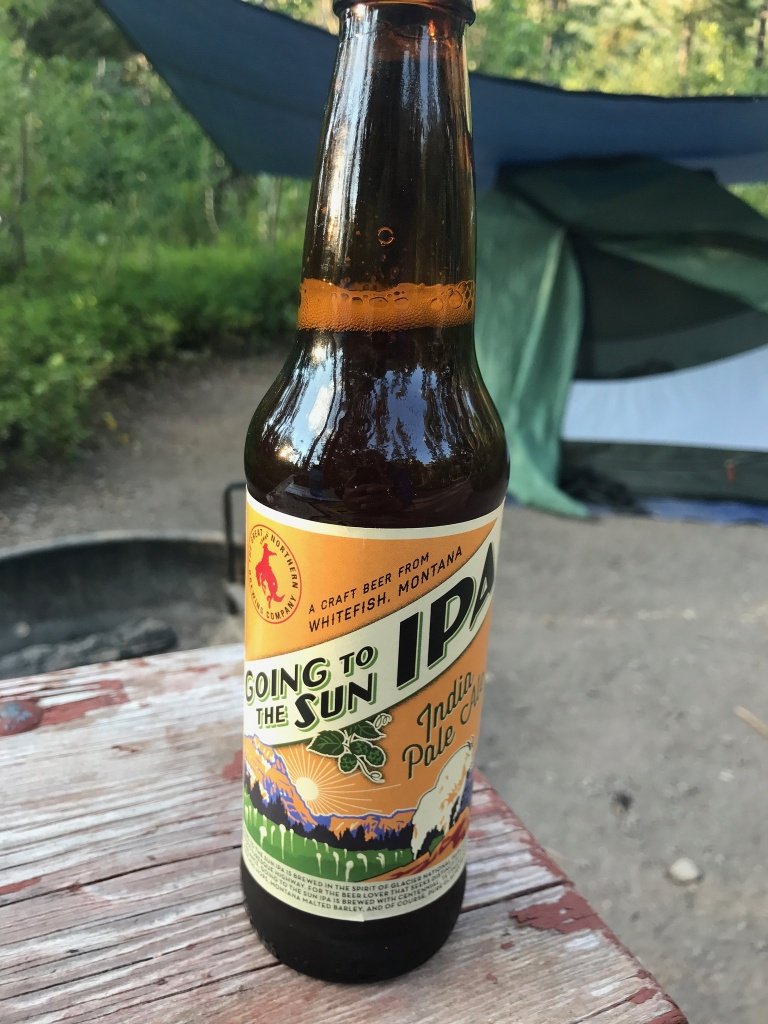}  
        &\includegraphics[height=0.15\textwidth,clip,trim=100 600 150 100]{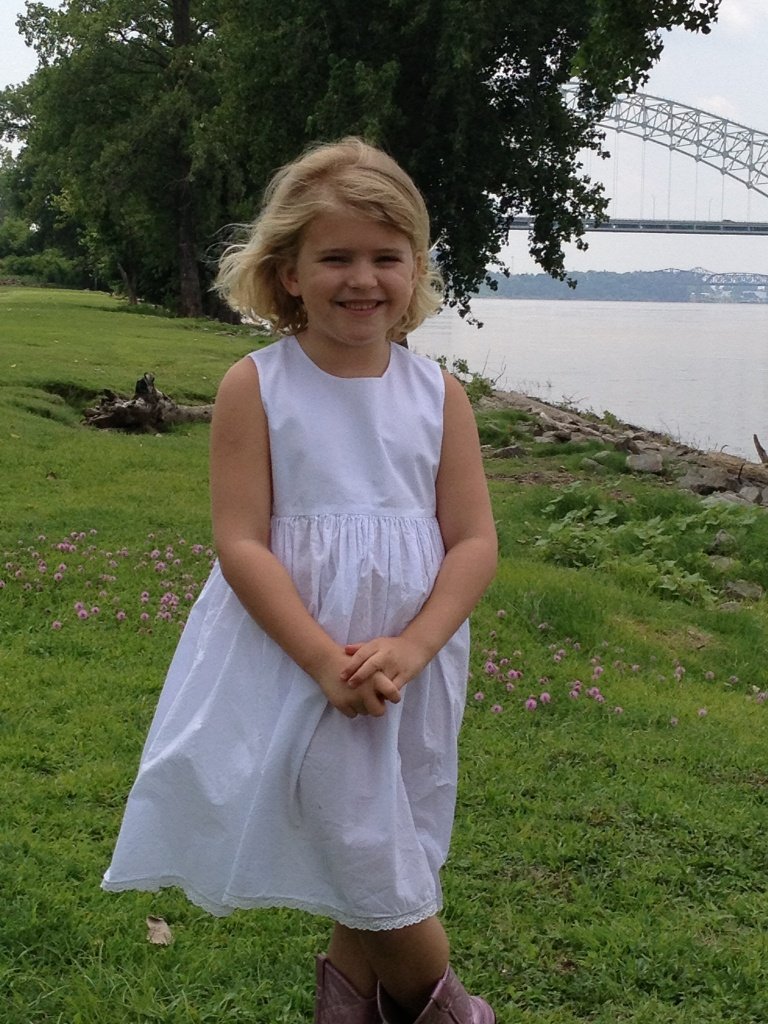}  
        &\includegraphics[height=0.15\textwidth,clip,trim=00 200 00 00]{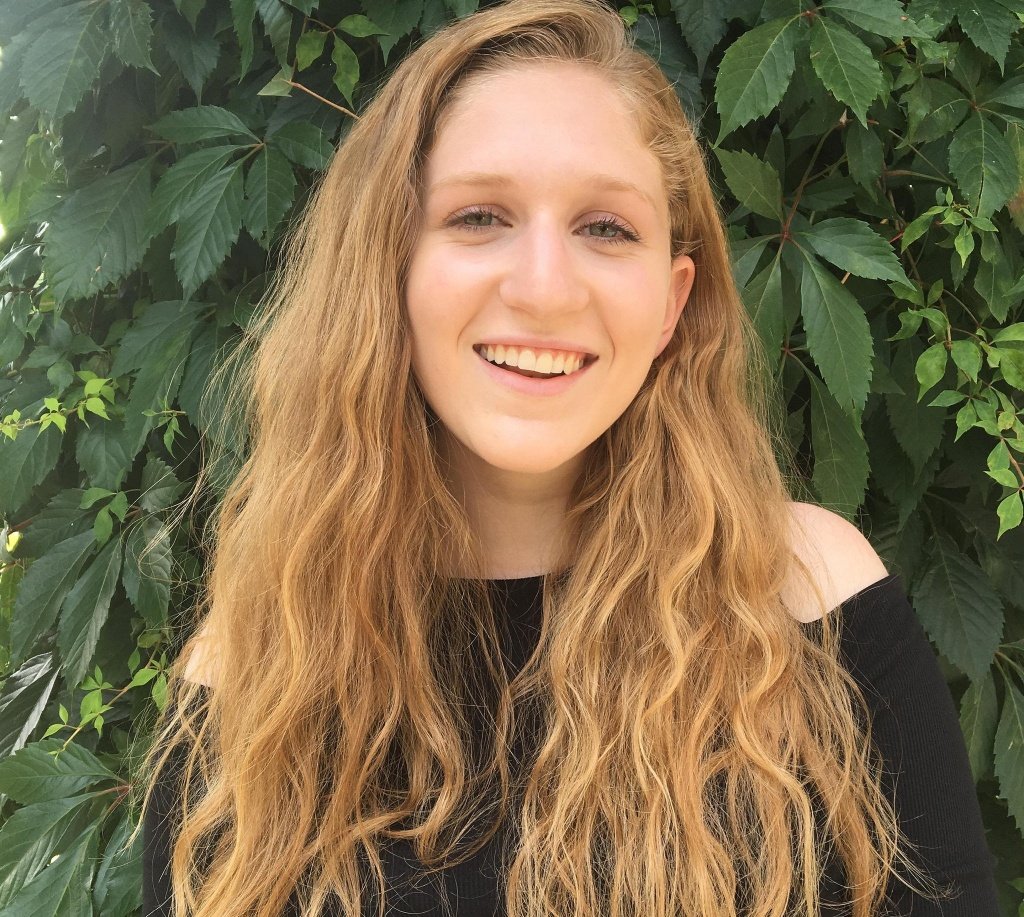} 
        \\
        \includegraphics[height=0.15\textwidth,clip,trim=50 400 00 125]{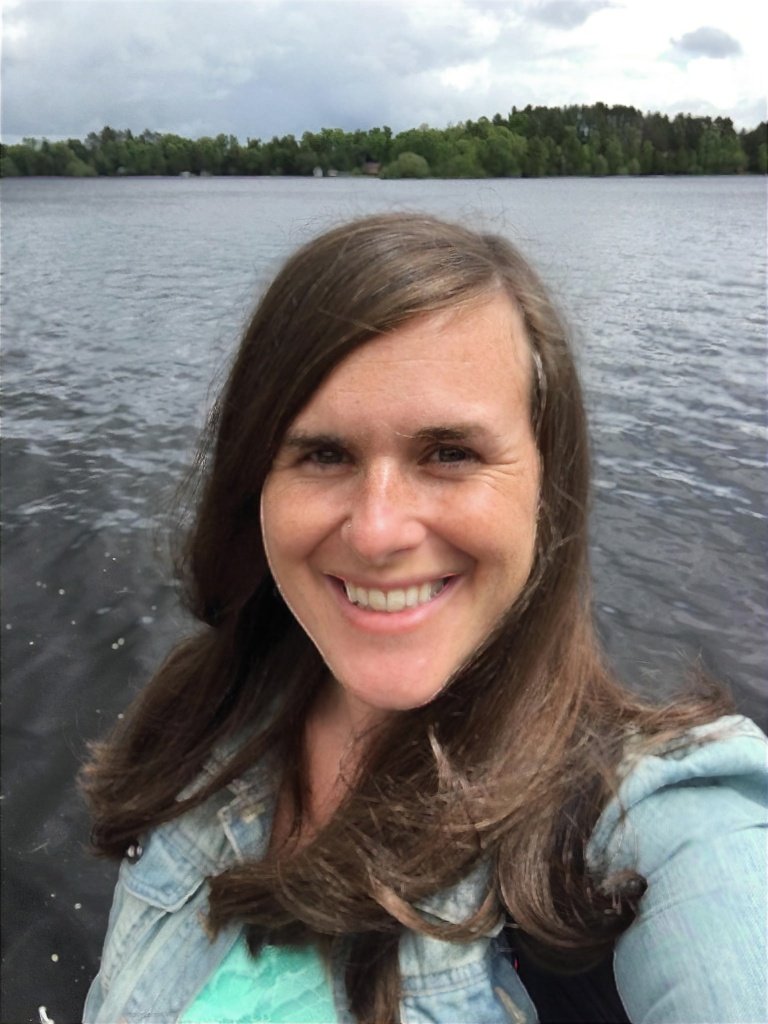}    
        &\includegraphics[height=0.15\textwidth,clip,trim=50 150 100 500]{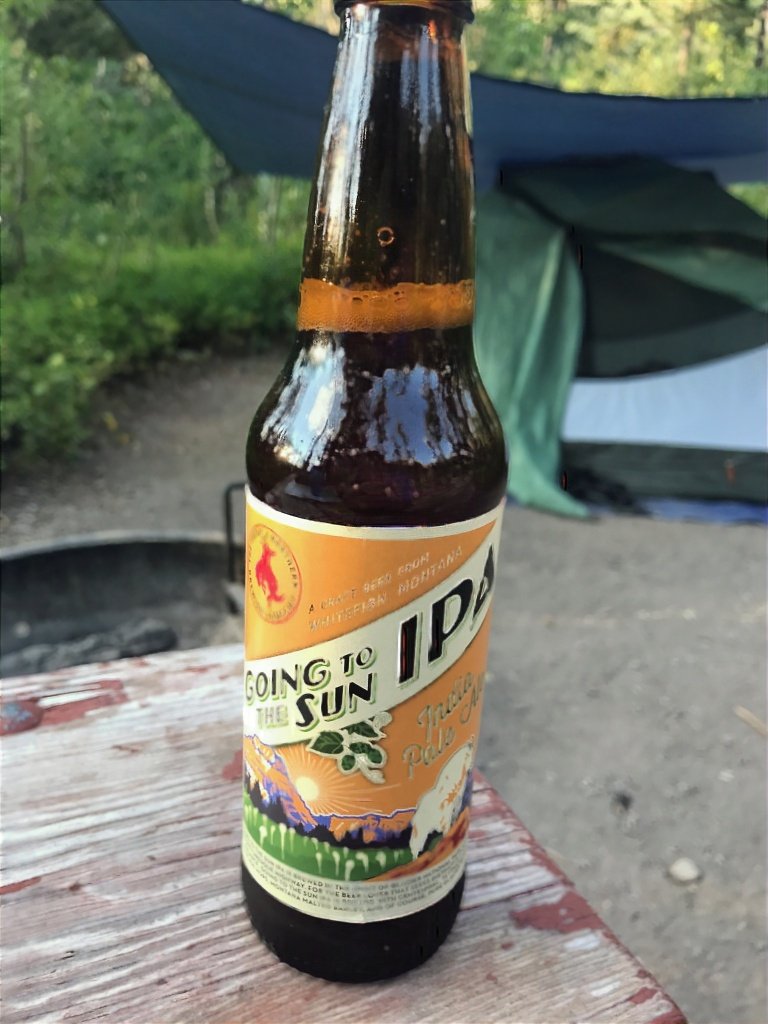}  
        &\includegraphics[height=0.15\textwidth,clip,trim=100 600 150 100]{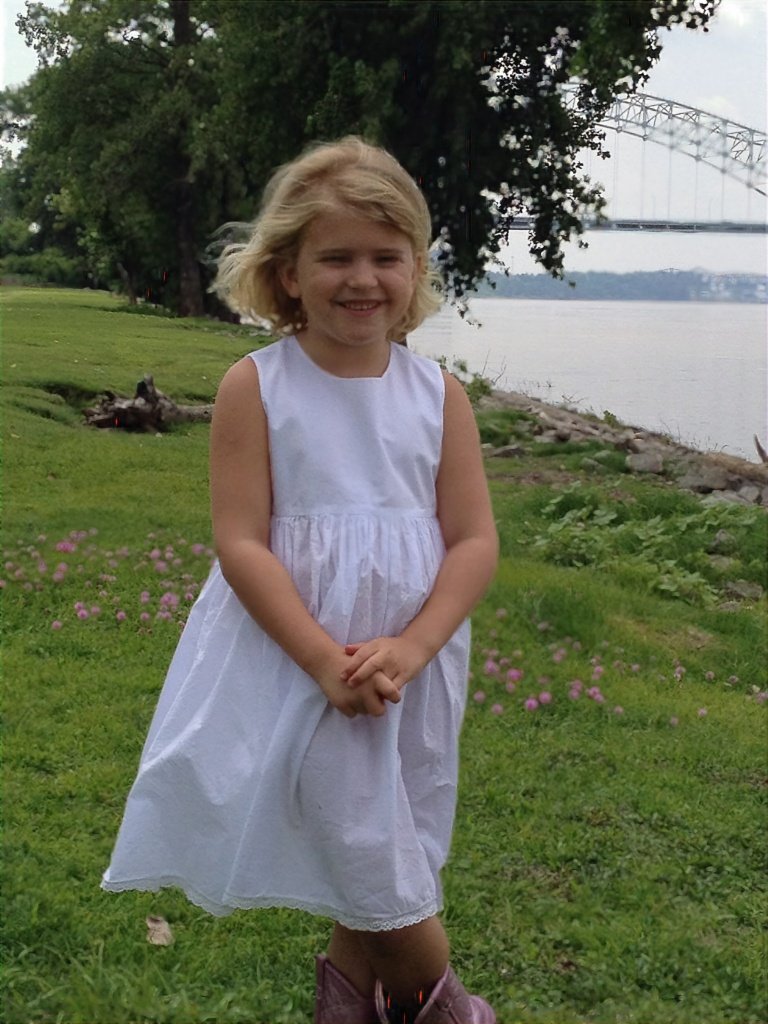}  
        &\includegraphics[height=0.15\textwidth,clip,trim=00 200 00 00]{{{figures/main/p2p/36800559341_fdc3ab17ee_k_kr_0.785_df_0.992_x_1063_y_512}}}  
        \\
        \includegraphics[height=0.15\textwidth,clip,trim=50 400 00 125]{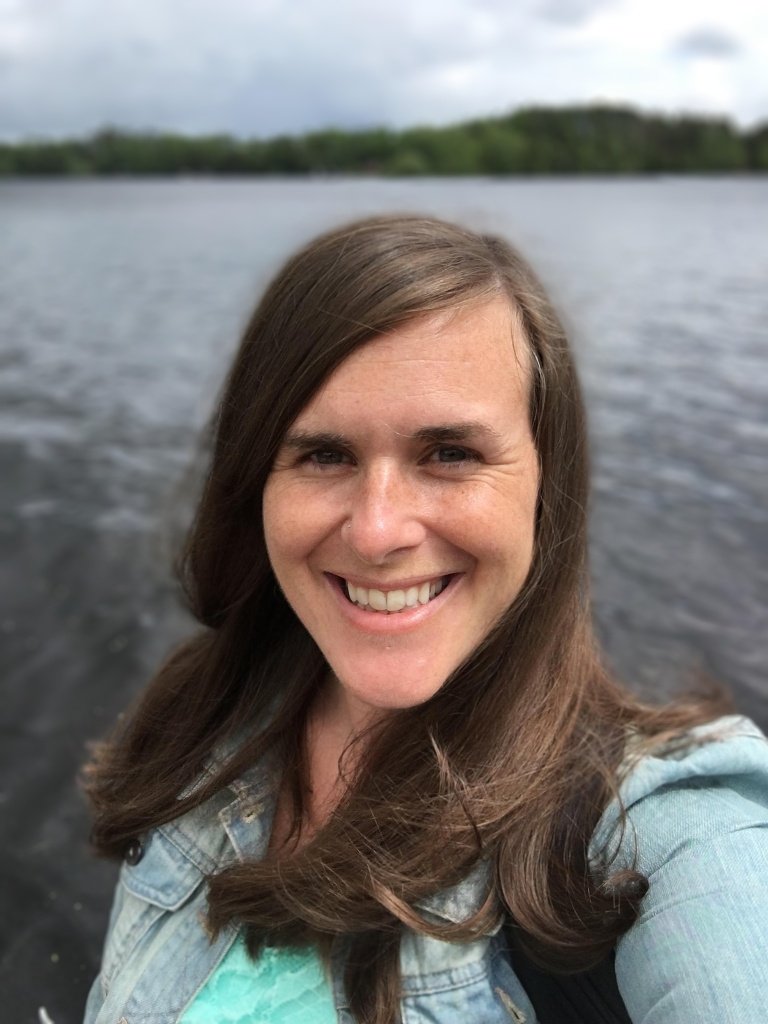}    
        &\includegraphics[height=0.15\textwidth,clip,trim=50 150 100 500]{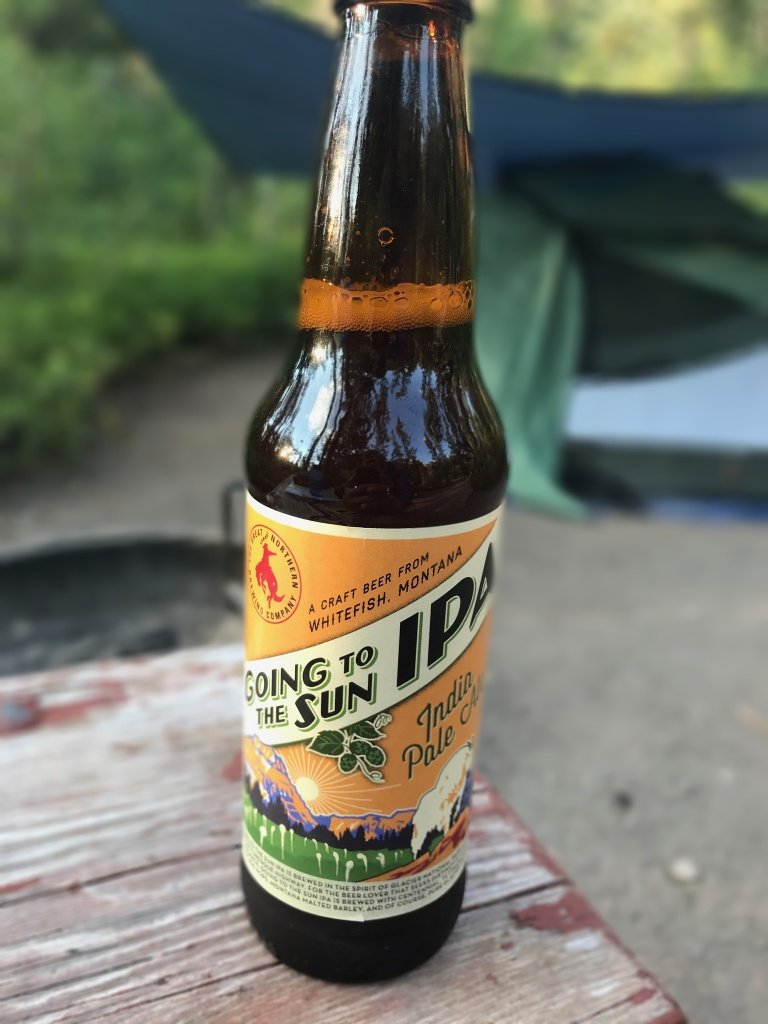}  
        &\includegraphics[height=0.15\textwidth,clip,trim=100 600 150 100]{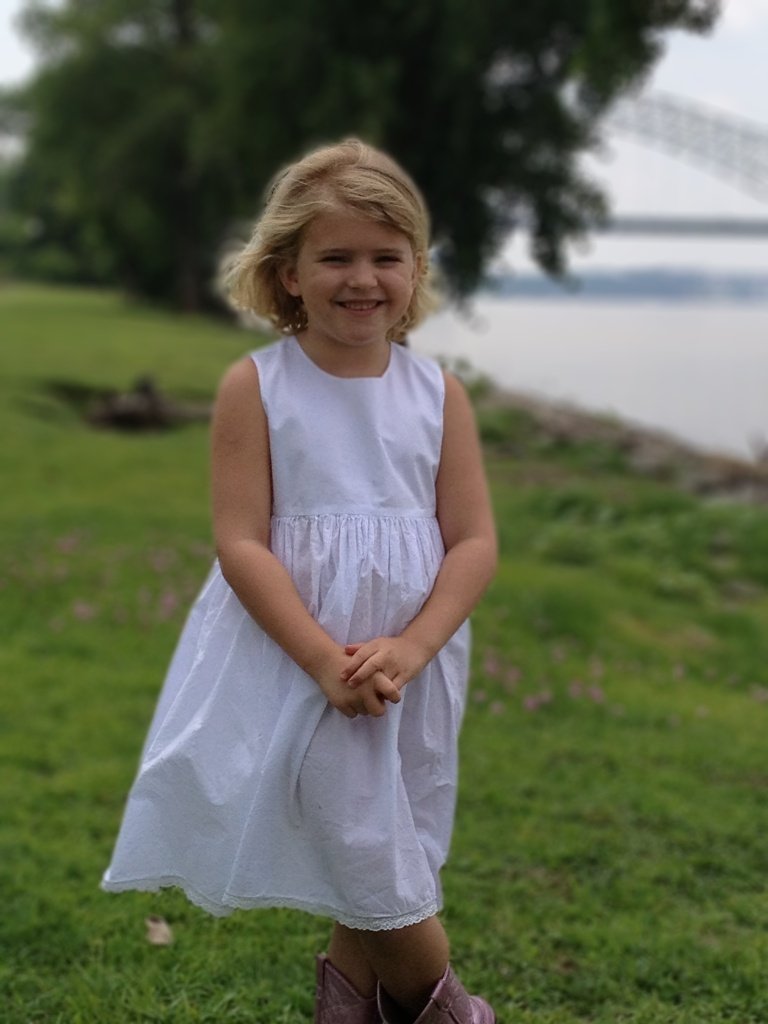}  
        &\includegraphics[height=0.15\textwidth,clip,trim=00 200 00 00]{{{figures/main/aperture/36800559341_fdc3ab17ee_k_kr_0.785_df_0.992_x_1063_y_512}}} 
        \\
        \includegraphics[height=0.15\textwidth,clip,trim=50 400 00 125]{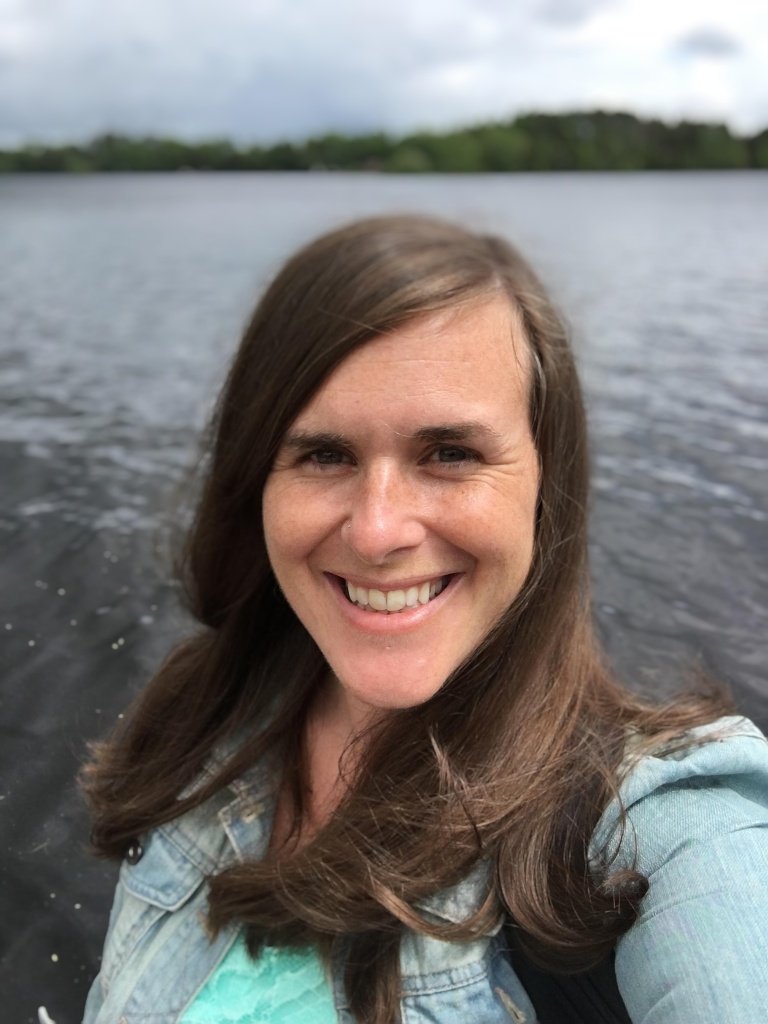}    
        &\includegraphics[height=0.15\textwidth,clip,trim=50 150 100 500]{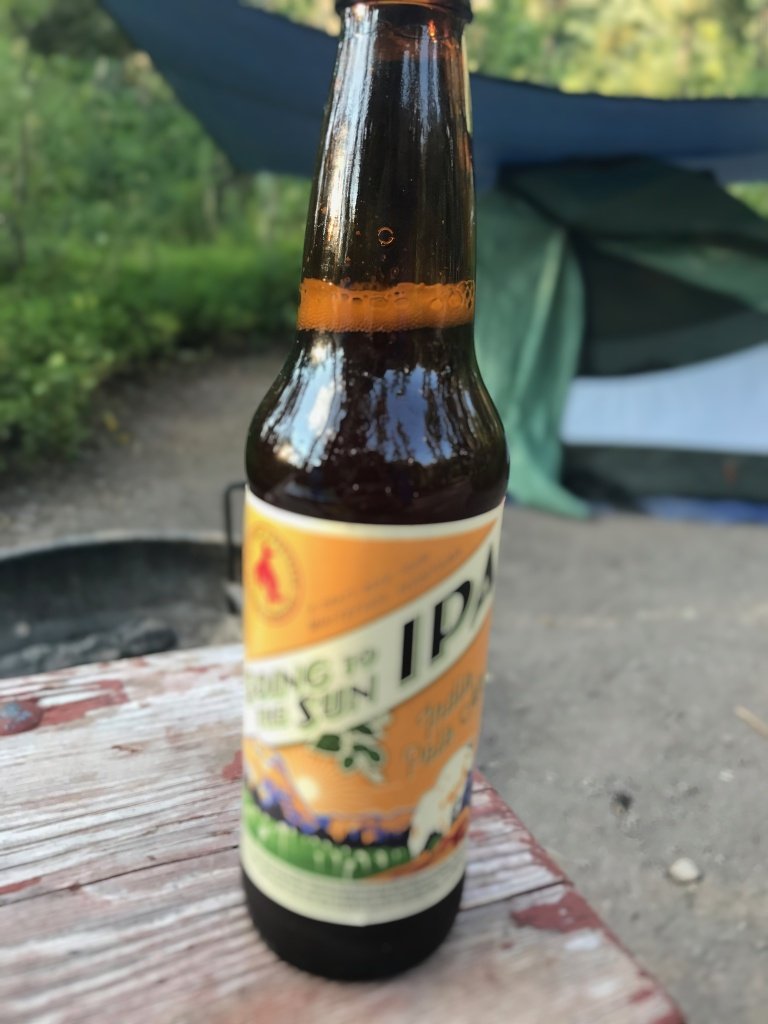}  
        &\includegraphics[height=0.15\textwidth,clip,trim=100 600 150 100]{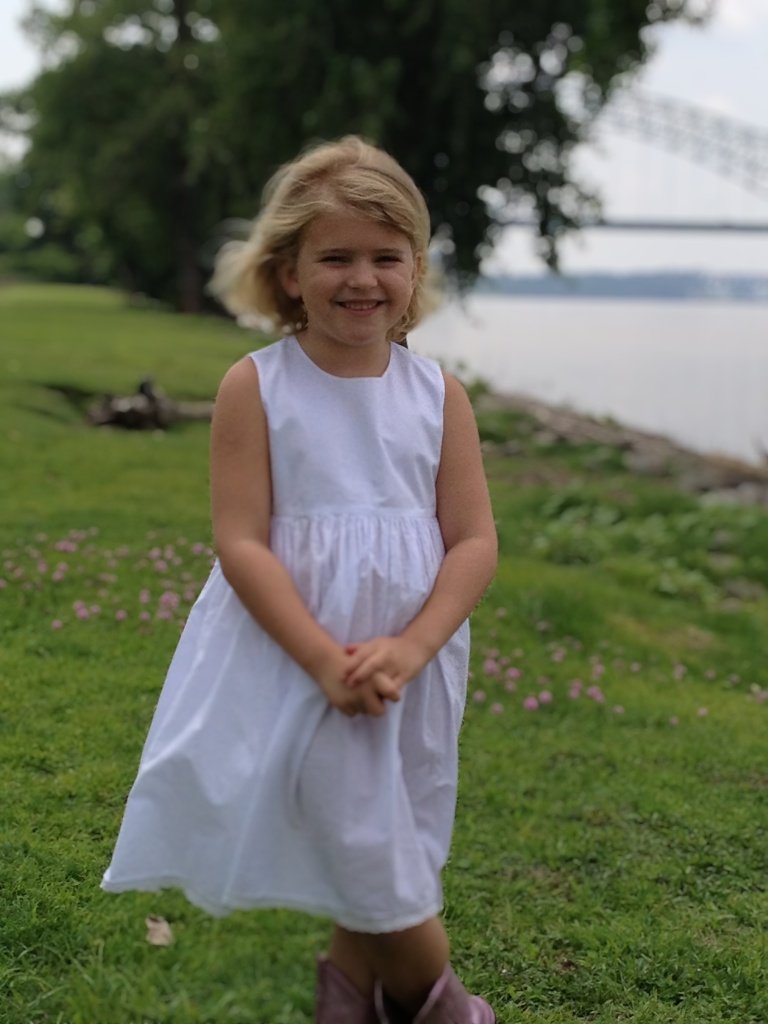}  
        &\includegraphics[height=0.15\textwidth,clip,trim=00 200 00 00]{{{figures/main/mega/36800559341_fdc3ab17ee_k_kr_0.785_df_0.992_x_1063_y_512}}} 
        \\
        \includegraphics[height=0.15\textwidth,clip,trim=50 400 00 125]{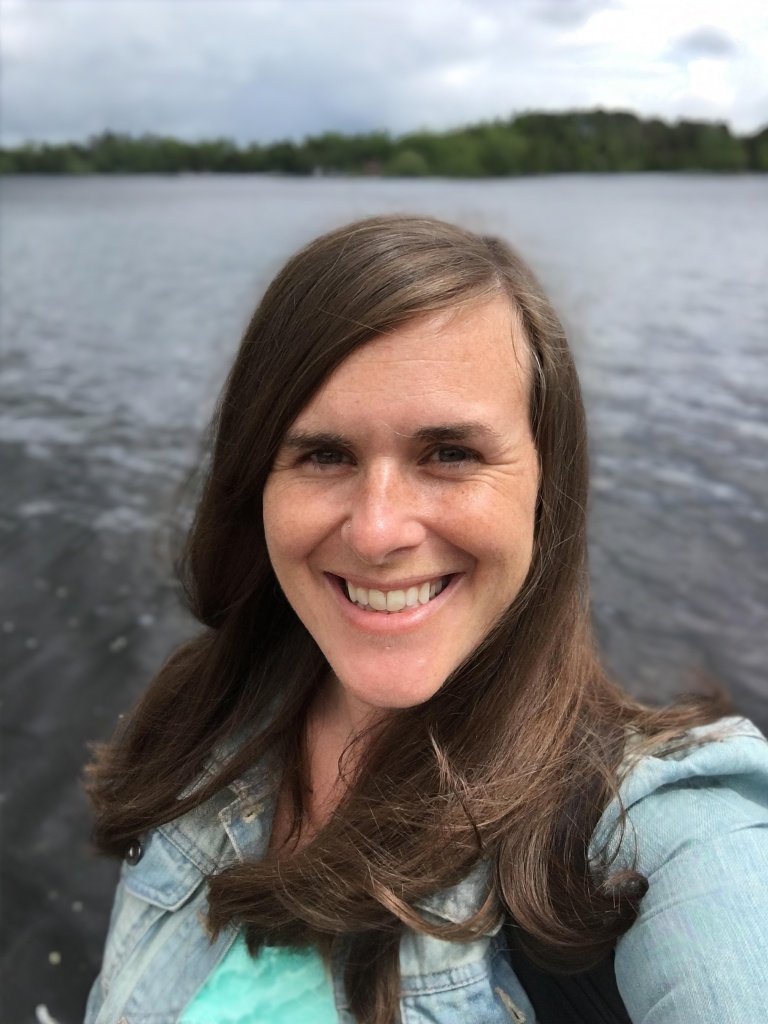}    
        &\includegraphics[height=0.15\textwidth,clip,trim=50 150 100 500]{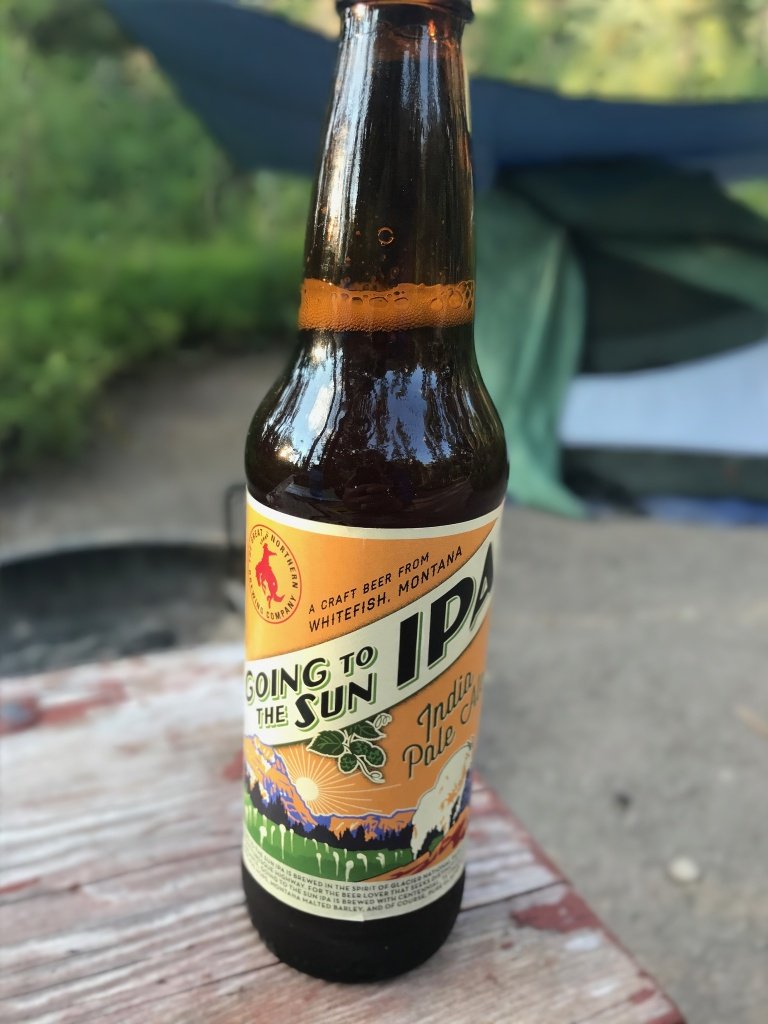}  
        &\includegraphics[height=0.15\textwidth,clip,trim=100 600 150 100]{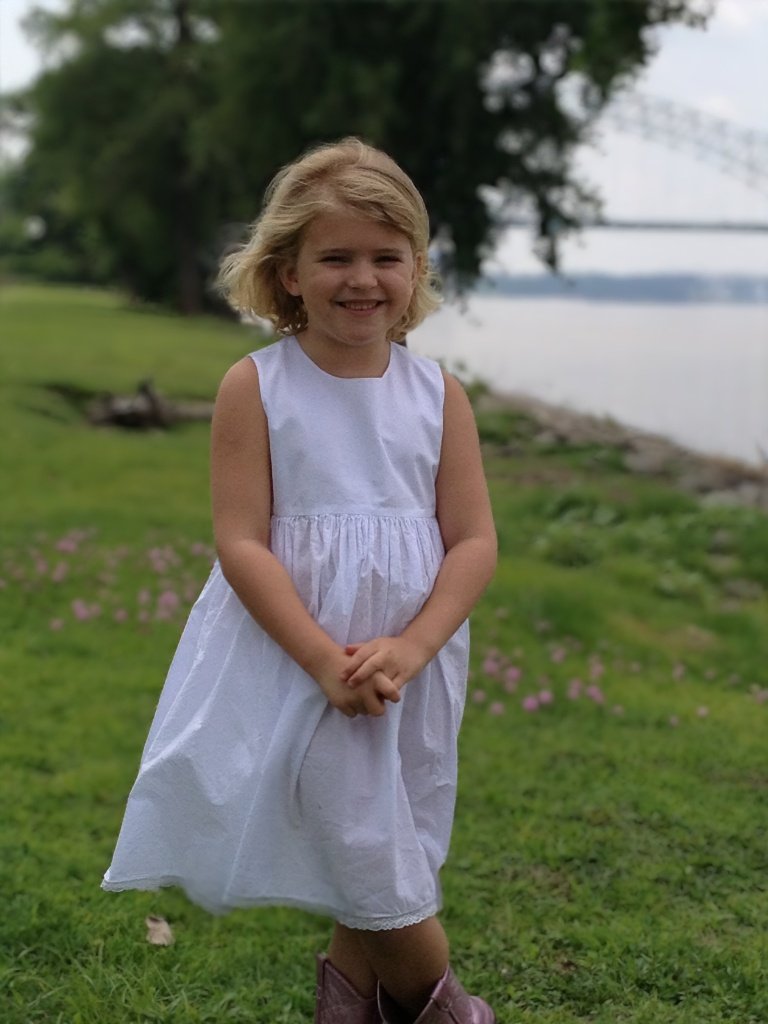}  
        &\includegraphics[height=0.15\textwidth,clip,trim=00 200 00 00]{{{figures/main/ours/36800559341_fdc3ab17ee_k_kr_0.785_df_0.992_x_1063_y_512}}}
	\end{tabular}
	\caption{Qualitative comparison of shallow DoF results on images from internet (Cropped). First row: input image; second row: results of direct regression; third row: results of Aperture; fourth row: results by MegaDepth~\shortcite{li2018megadepth}+\cite{yang2016virtual}; last row: our results. Please zoom-in to see details. \textit{Photo credits: Erin Page, daveynin, mrchrishill and Jewish Women's Archive.}}
	\label{fig:dof_web}
\end{figure*}

We first evaluate the performance of our depth prediction module. 
We compare our approach with two state-of-art methods including \cite{laina2016deeper} and MegaDepth\cite{li2018megadepth}, which are trained to predict absolute depth values on the NYU v2 data set~\cite{Silberman2012nyuv2} and the MegaDepth data set~\cite{li2018megadepth}, respectively. 
In addition, we also implement a baseline method named Aperture, which has the same network architecture as ours and is trained with supervision only on DoF images, similar to the method in~\cite{pratul2018}. 
For a fairer comparison, we follow \cite{pratul2018} and fit a 5-knot linear spline to minimize the squared error of each prediction with respect to the ground truth.
This is needed to correct for scale differences. 
Especially for aperture supervision, which can produce an inverted depth map. 

\final{Fig.~\ref{fig:depth_test} and \ref{fig:depth_web} illustrate qualitative comparisons on the iPhone test set and random internet images, respectively. 
It can be observed that the models trained on existing depth datasets struggle to generalize to both our test set and internet images, while our method trained on the proposed depth data set achieves perceptually more accurate results across the test images. 
Compared with the Aperture baseline, which is our model trained end-to-end with supervision on shallow DoF images, we observe that our method more clearly delineates object boundaries.}

\begin{table}
    \centering
    \small
	\caption{\label{tab:depth}Comparison of depth prediction methods on our iPhone depth test set in terms of mean absolute error (MAE). \final{Results in parentheses are achieved after training the corresponding methods on our iPhone depth training set, and the Aperture model is trained end-to-end using shallow DoF supervision.}}
    \resizebox{.99\linewidth}{!}{
	\setlength{\tabcolsep}{3pt}
		\begin{tabular}{c  c  c  c  c}
			\toprule[1pt]
			Method & \cite{laina2016deeper} & MegaDepth~\shortcite{li2018megadepth}& Aperture & Ours \\
			\midrule
			MAE & 0.199 \final{(0.138)} & 0.182 \final{(0.142)} & 0.146 & 0.116 \\
			\bottomrule[1pt]
		\end{tabular}
	}
	\vspace{-4mm}
\end{table}

\final{
A comparison of results are reported in terms of mean absolute error (MAE) in Tab.~\ref{tab:depth}. We report the performance of \cite{laina2016deeper} and MegaDepth~\cite{li2018megadepth} trained on their proposed datasets and tested on our iPhone dataset, as well as trained on our same training set.}

\begin{figure*}[t]
	\centering
	\small
	\resizebox{.99\linewidth}{!}{
	\tabcolsep0.3pt \renewcommand{\arraystretch}{0.5}
	\begin{tabular}{cccc}
		\includegraphics[width=0.23\textwidth,clip,trim=030 110 150 50]{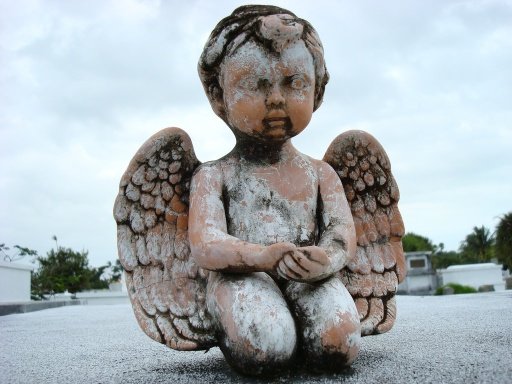}  &
		\includegraphics[width=0.23\textwidth,clip,trim=030 110 150 50]{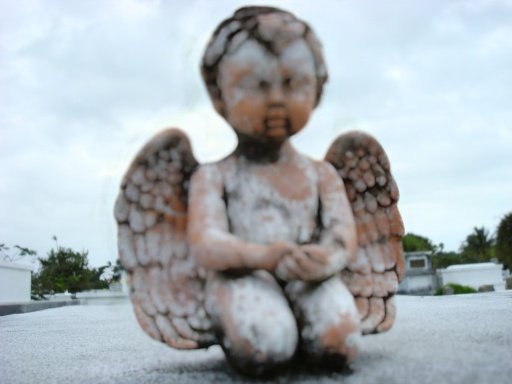}&
		\includegraphics[width=0.23\textwidth,clip,trim=030 110 150 50]{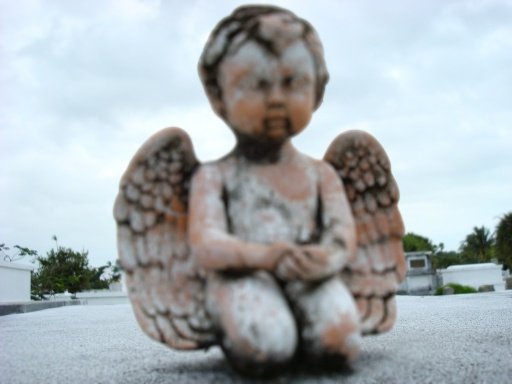}&
		\includegraphics[width=0.23\textwidth,clip,trim=030 110 150 50]{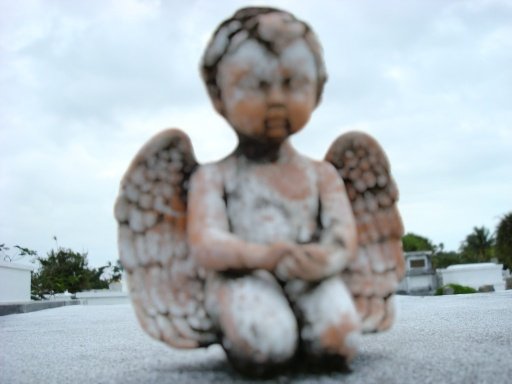}\\
		\includegraphics[width=0.23\textwidth,clip,trim=130 590 100 65]{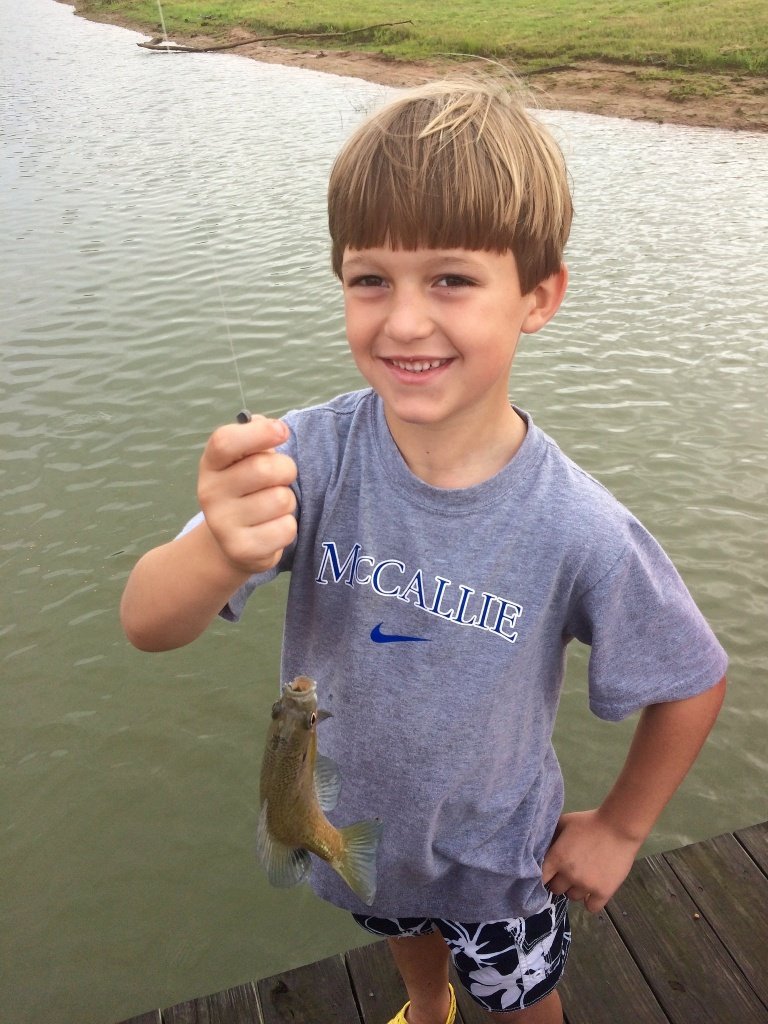}  &
		\includegraphics[width=0.23\textwidth,clip,trim=130 590 100 65]{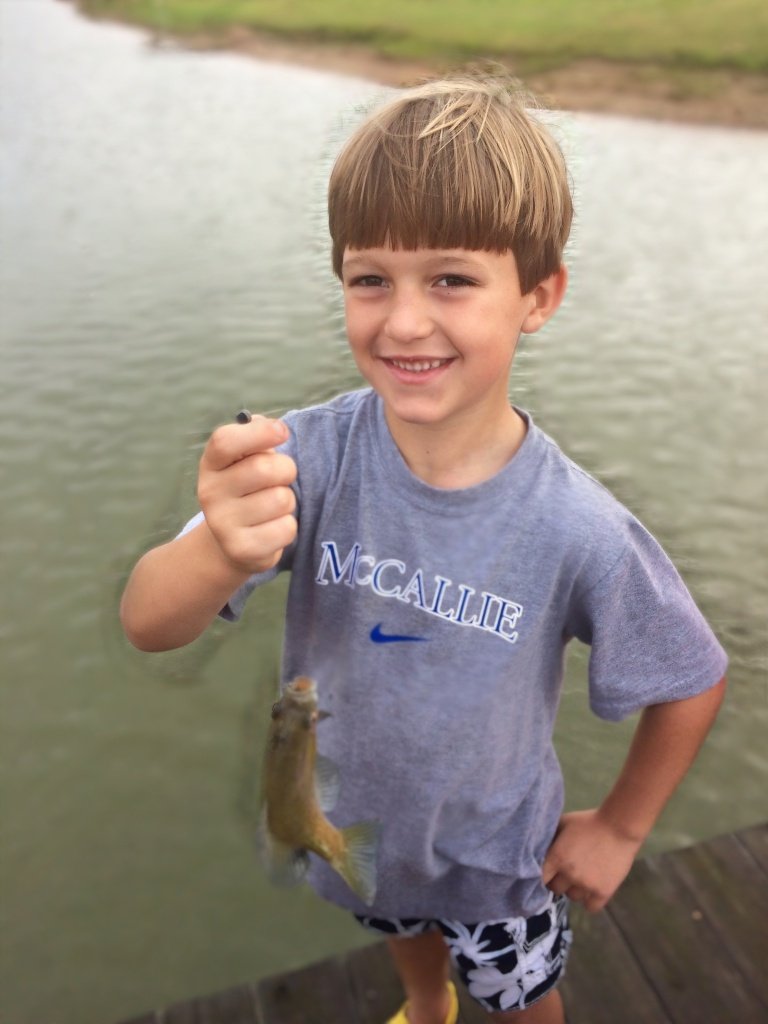}&
		\includegraphics[width=0.23\textwidth,clip,trim=130 590 100 65]{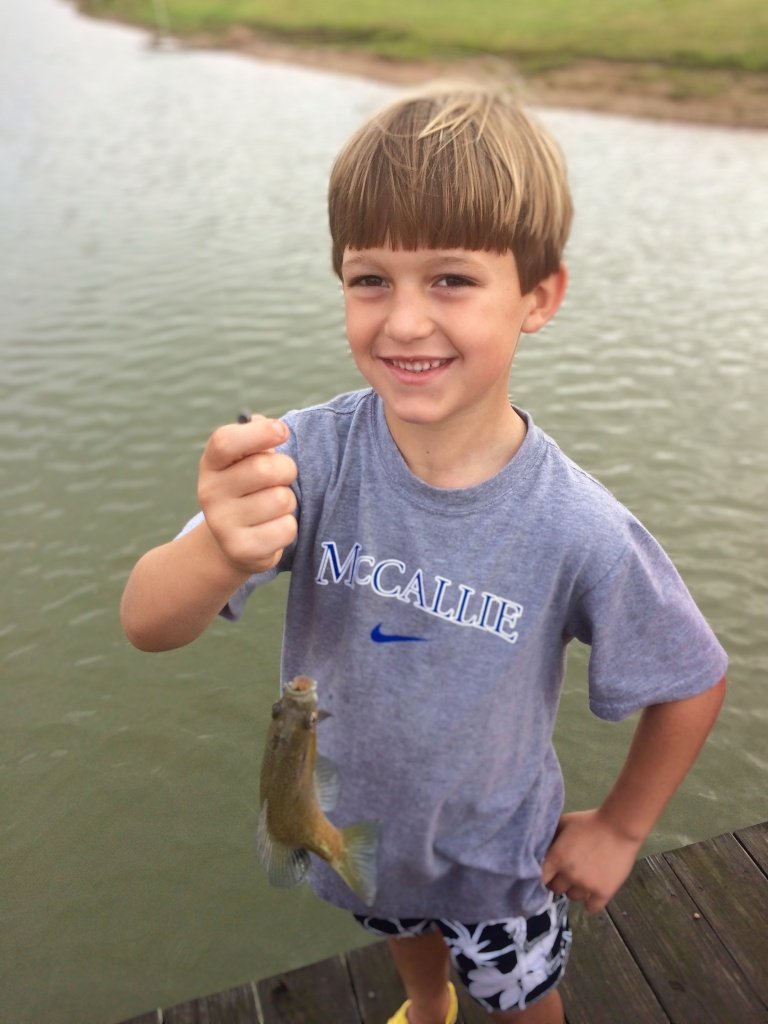}&
		\includegraphics[width=0.23\textwidth,clip,trim=130 590 100 65]{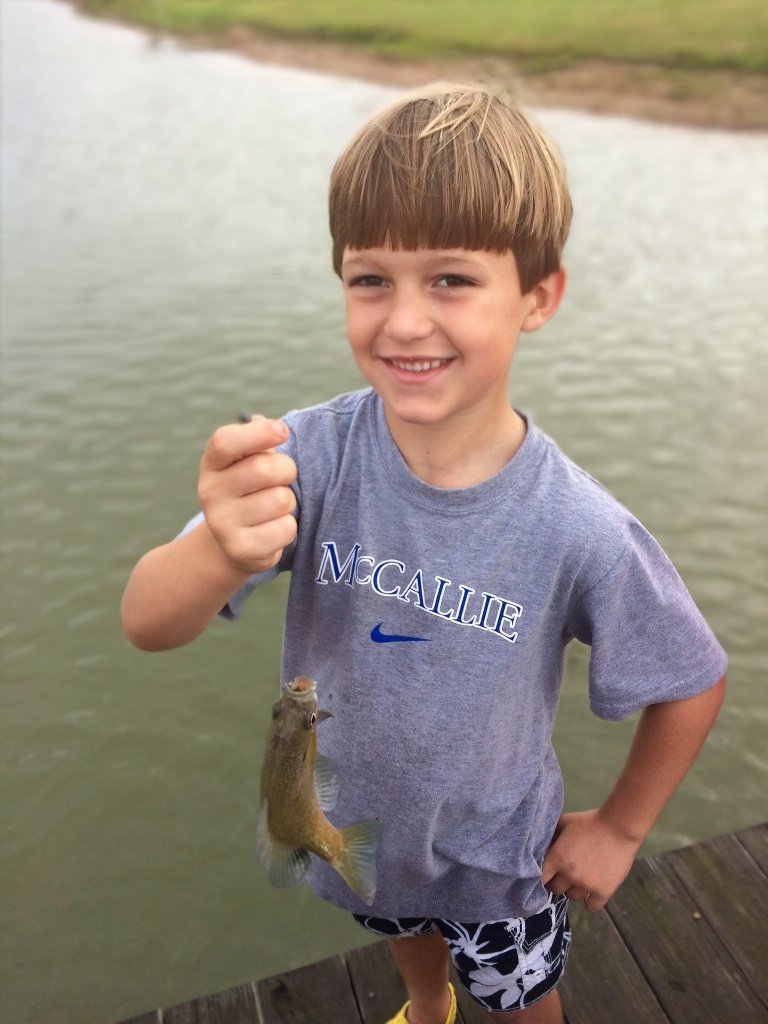}\\
		\\
		Input & Aperture + Our Lens Blur & Our Depth + ~\cite{yang2016virtual} & Ours
	\end{tabular}
	}
    \caption{Qualitative comparison of shallow DoF results (Cropped). \textit{Photo credits: Christine and mrchrishill.}}
	\label{fig:ablative}
\end{figure*}

\subsection{Evaluation on DoF results}

To evaluate the quality of our generated shallow DoF results, we compare to a strong baseline consisting of a state-of-the-art single image depth prediction method, MegaDepth~\cite{li2018megadepth}, and an approximate ray tracing based shallow DoF effect method from \cite{yang2016virtual}. \final{For a fair comparison, we use the model of MegaDepth trained on our iPhone depth data set.
}
We also compare to direct DoF image regression (without explicitly inferring the depth map), using a state-of-the-art image-to-image translation method~\cite{Isola2017ImagetoImageTW}, which has previously shown successful in blur magnification to generate shallow DoF images. 
To train the image-to-image translation model, we feed the focal point, desired aperture size and the source image to the network, and ask the network to predict the resulting shallow DoF image.
\final{In addition, we have also compared to the Aperture baseline (See Section~\ref{sec:depth}), which combines our depth prediction network with a differentiable compositional DoF rendering approach~\cite{pratul2018}, and is trained in an end-to-end manner with DoF supervision.} 
We quantitatively evaluate all compared methods on the iPhone test set, with ground truth rendered by \cite{yang2016virtual} based on ground truth iPhone depth maps. The comparison in terms of PSNR and SSIM are shown in Tab.~\ref{tab:DoF}, and qualitative results on internet images are illustrated in Fig.~\ref{fig:dof_web}.  
Our method has achieved superior performance against the other baselines, both quantitatively and qualitatively. 
This is in part due to the multi-task training on a segmentation task, and our new iPhone Depth data set with more diverse scenarios.
The direct regression method often fails to render shallow DoF effects and suffers from severe artifacts, while results generated by Aperture and the combination of two state-of-the-art prior works are more perceptually plausible, but have artifacts in the predicted depth discontinuities, which affect the shallow DoF rendering. Note that the quantitative evaluation on iPhone test set would be inevitably affected by the artifacts in the rendered ground truth as mentioned in Section \ref{sec:iphonedata}. Therefore, we complement the evaluation with a user study on internet photos (Section \ref{sec:userstudy}).

\begin{table}\centering
	\small
	\caption{\label{tab:DoF}Comparison of DoF results on iPhone test set.}
    \resizebox{.99\linewidth}{!}{
	\setlength{\tabcolsep}{3pt}
		\begin{tabular}{c | c | c | c | c }
			\toprule[1pt]
			Method & Direct Regression &Aperture & MegaDepth\shortcite{li2018megadepth} + \cite{yang2016virtual} & Ours \\
			\hline
			PSNR & 25.84 &26.88 &26.67 &28.235 \\
			SSIM & 0.815 &0.844 &0.836 &0.908 \\
			\bottomrule[1pt]
		\end{tabular}
	}
\end{table}

\begin{figure}
    \centering
    \begin{tabular}{cc}
    \includegraphics[width=.5\linewidth, clip, trim=50 0 100 0]{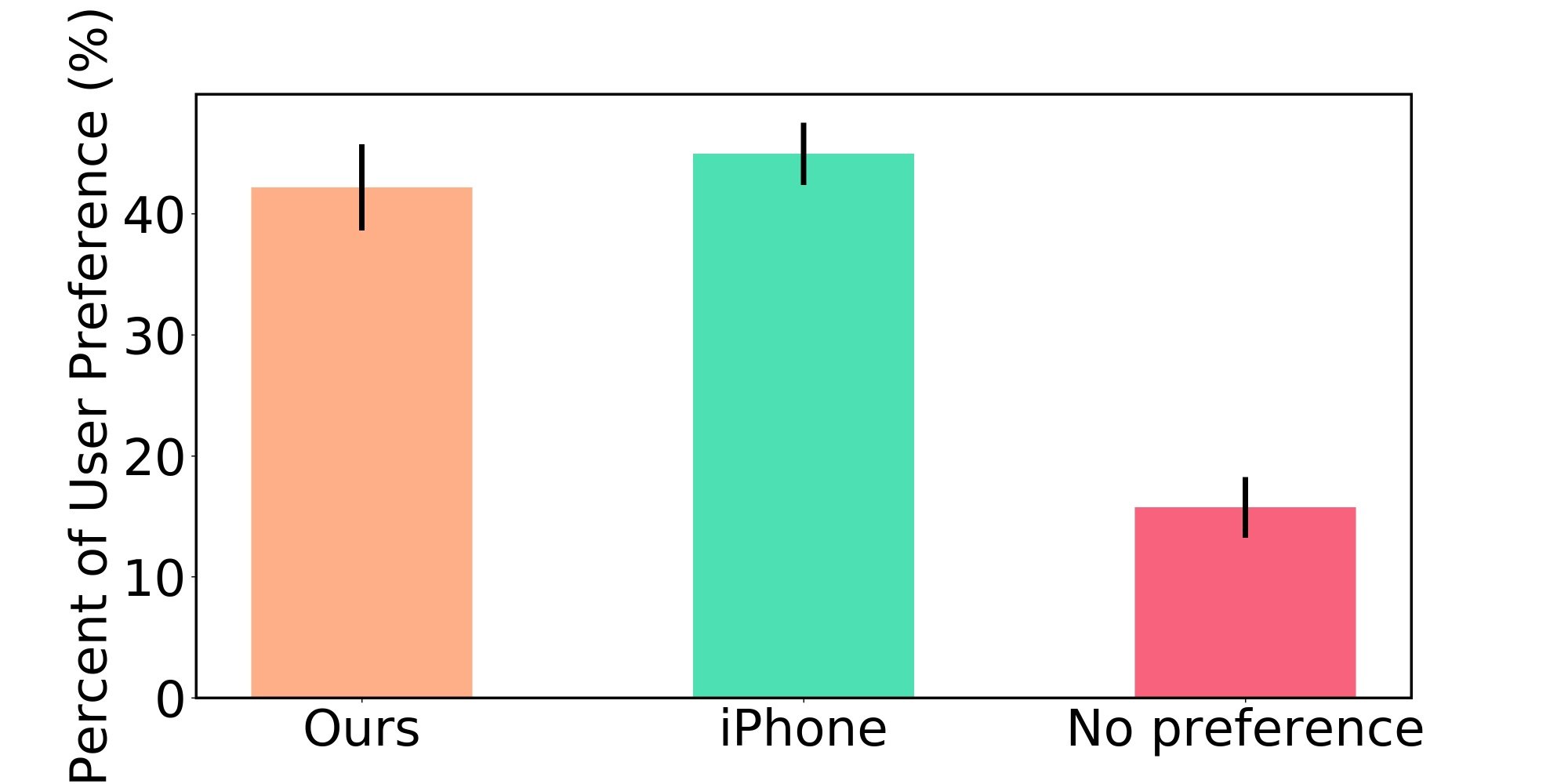} &
    \includegraphics[width=.5\linewidth,  clip, trim=50 0 100 0]{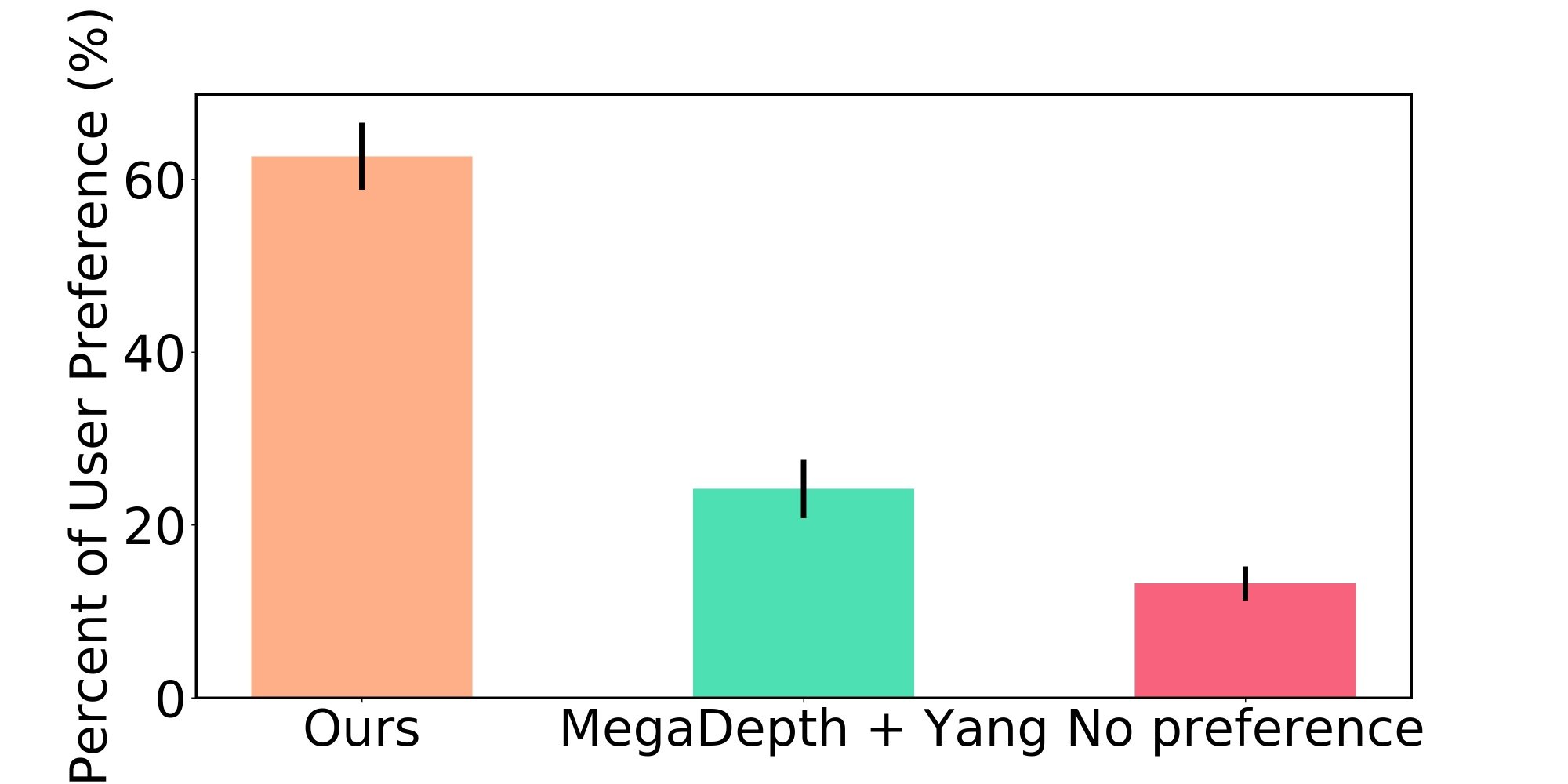} \\
    \end{tabular}
    \caption{The results of our user study. Our method is compared to iPhone portrait mode images (left), and the ray-tracing based approach~\cite{yang2016virtual} (right). Note that the both the iPhone and~\cite{yang2016virtual} are computed with measured depth, whereas our method uses only a single image. Error bars show 95\% confidence intervals.}
    \label{fig:userstudy}
\end{figure}

\subsection{Ablative Study}\label{sec:ablative}
To understand the contribution of each module, we compare our final method to another two baselines. 
In the first baseline, we replace the depth prediction network with the Aperture method (Section~\ref{sec:depth}) and use our lens blur and guided upsampling to render shallow DoF, which we refer to as Aperture + Our Lens Blur. 
In the second baseline, we combine our depth prediction network and the DoF rendering method \cite{yang2016virtual}, which we refer to as Our Depth + \cite{yang2016virtual}.
As shown in Fig.~\ref{fig:ablative}, Aperture + Our Lens Blur performs poorly mainly due to inaccurate depth predictions, suggesting that that depth supervision is still essential for high quality results. 
With more accurate depth prediction, Our Depth + \cite{yang2016virtual} method works well, however, it still suffers from artifacts at depth discontinuities on the foreground boundary. 
In comparison, our lens blur method is trained on an artifact-free data set and takes both depth and images as input, which allows it to compensate subtle errors in depth prediction, and avoid common artifacts of image-based shallow DoF approximations. 

\begin{figure}
    \centering
    \includegraphics[width=0.5 \textwidth]{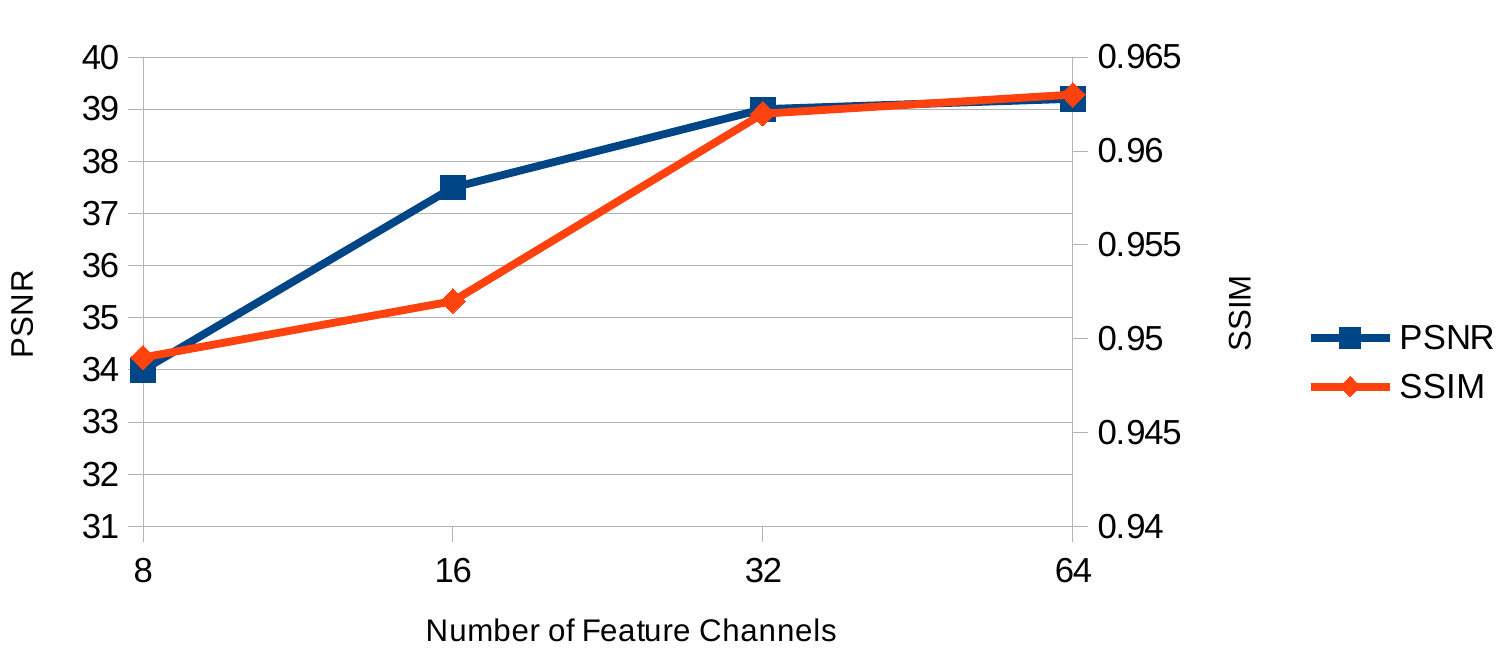}
    \caption{ PSNR and SSIM of our method with respect to numbers of feature channels on the test set of the synthetic shallow DoF dataset}
    \label{fig:channel}
\end{figure}

Recall that our lens blur module is performed in a learned feature space. 
To investigate the effect of feature dimension on the final results, we evaluate the performance of our method with respect to different numbers of feature channels on the test set of the synthetic shallow DoF dataset. 
As we see in Fig.~\ref{fig:channel}, the quality of the rendered results increases with respect to the channel number and begins to saturate at 32 channels.
For both efficiency and effectiveness, we adopt 32 channels in our final models, which amounts to total of 12.5\% of the kernel size represented, showing that we achieve a large memory savings with almost no quality loss by operating in a learned feature space. 

\begin{table}\centering
    \small
	\caption{\label{tab:ablative}\final{Comparison of DoF results on iPhone test set. RealDepth and SynDepth denotes the depth prediction modules which are trained on the iPhone depth dataset and synthetic data set, respectively. GT, Pred, and Corrupted indicate that the lens blur and guided upsampling modules are trained with ground truth, predicted and corrupted depth, respectively.}}
    \resizebox{.99\linewidth}{!}{
	\setlength{\tabcolsep}{3pt}
		\begin{tabular}{c | c | c | c | c }
			\toprule[1pt]
			
			Method & Baseline1 &Baseline2 &Baseline3 & Ours \\
			\hline
			Depth Module &RealDepth &RealDepth &SynDepth &RealDepth \\
			Training Input &GT &Pred &Pred &Corrupted\\ 
			PSNR & 28.038 & 28.237 &22.380 &28.235 \\
			\bottomrule[1pt]
		\end{tabular}
	}
\end{table}


\final{To analyze the impact of learning lens blur and guided upsampling from corrupted depth maps (Section~\ref{sec:syn}), we compare our method with three baselines.
The depth prediction modules of the first two baselines and our method are the same and they are trained on the iPhone depth data (denoted as the RealDepth models).
However, for training the lens blur and guided upsampling modules, the first baseline uses the ground truth synthetic depth, and the second baseline uses the depth generated by a depth prediction model trained on the synthetic data, denoted as the SynDepth model.
As shown in Tab.~\ref{tab:ablative}, our method trained with corrupted depth achieves similar performance to the second one trained with predicted depth, which is superior to training with ground truth depth. This indicates that it can be sub-optimal to train the lens blur and guided upsampling modules on the ground truth depth because the networks can be sensitive to the errors made by the depth prediction module. The second baseline uses predicted depth to achieve more robustness, but it requires training an additional SynDepth model, because the RealDepth model completely fails on the synthetic depth data.
Furthermore, we evaluate a third baseline by replacing the depth prediction module of the second baseline with the SynDepth model. The comparison between the second and the third baselines suggests that the SynDepth model trained on synthetic depth fails to generalize to real scenes, leading to low quality of the final DoF results. 
As a matter of fact, the MAE of the SynDepth on the iPhone test set is 0.30, which is significantly worse than other competitors listed in Tab.~\ref{tab:depth}. This further confirms the necessity of training depth prediction and lens blur modules on separate data sets.
}

\ignore{To analyze the impact of learning lens blur and guided upsampling from corrupted depth maps (Section~\ref{sec:syn}), we compare our method with three baselines. 
The first is trained on ground truth depth maps from the synthetic data set, while the second and the third are trained with input depth maps generated by a depth prediction module trained on the synthetic data. Then, at test time the first and second baselines use the depth prediction module trained on the iPhone depth dataset while the third baseline uses the depth prediction module trained on the synthetic data set. 
Quantitative results in terms of PSNR are reported in Tab.~\ref{tab:ablative}. By comparing our method with the first two baselines, we can observe that our method trained with corrupted depth achieves similar performance to the second one trained with predicted depth, which is superior to training with ground truth depth. Meanwhile, our method does not require training an additional depth prediction module, while the second baseline does. The comparison between the second and the third baselines suggests that the depth prediction module trained on synthetic depth fails to generalize to real scenes, leading to low quality of the final DoF results. 
As a matter of fact, the MAE of the depth prediction module trained on synthetic depth is 0.30, which is significantly worse than other competitors listed in Tab.~\ref{tab:depth}, further confirming the necessity of training depth prediction and lens blur modules on separate data sets.
}

\begin{figure}
	\centering
	\small
	\tabcolsep0.3pt \renewcommand{\arraystretch}{0.5}
	\begin{tabular}{ccc}
		\includegraphics[width=0.3\linewidth,clip,trim=0 125 0 125]{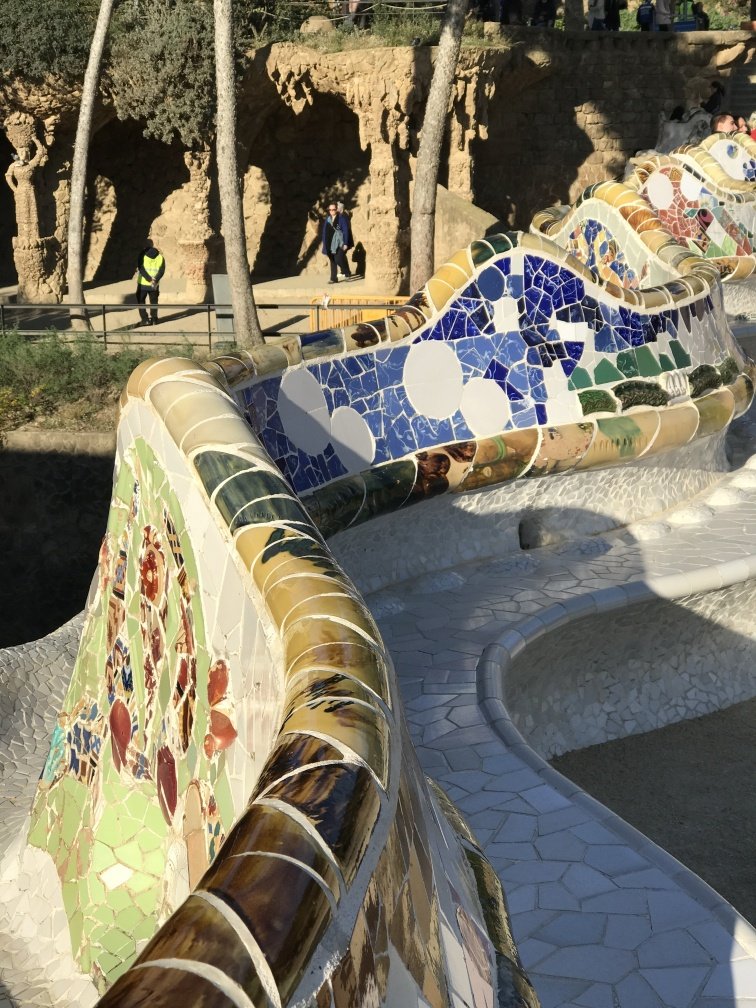}  &
		\includegraphics[width=0.3\linewidth,clip,trim=0 125 0 125]{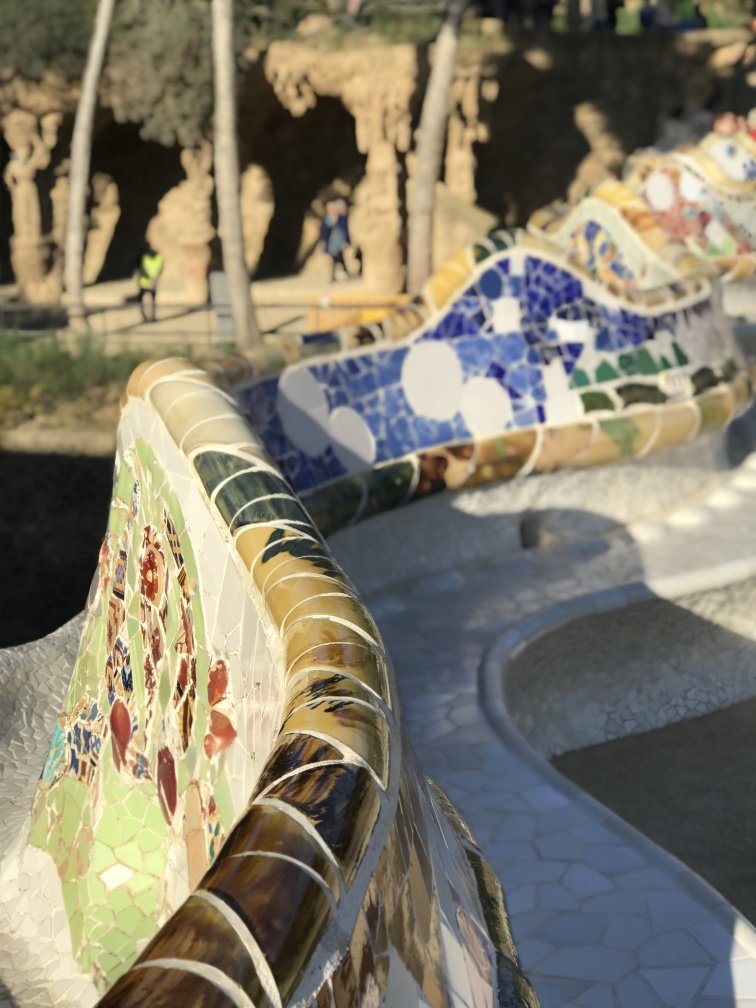} &
		\includegraphics[width=0.3\linewidth,clip,trim=0 125 0 125]{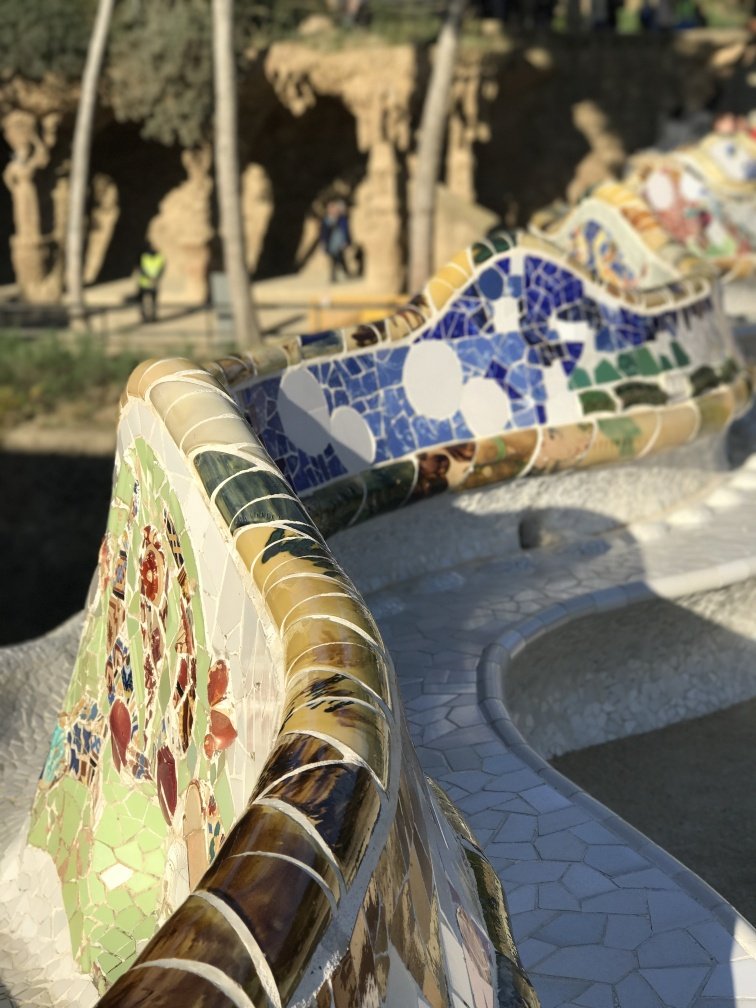}\\
		\includegraphics[width=0.3\linewidth,clip,trim=0 125 0 125]{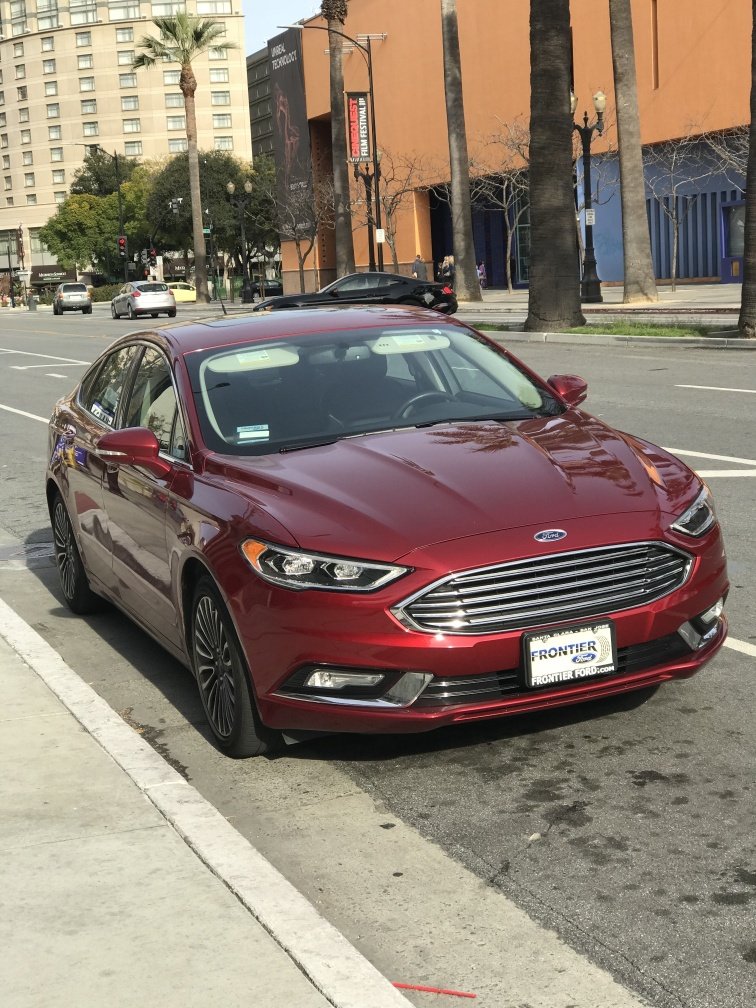}  &
		\includegraphics[width=0.3\linewidth,clip,trim=0 125 0 125]{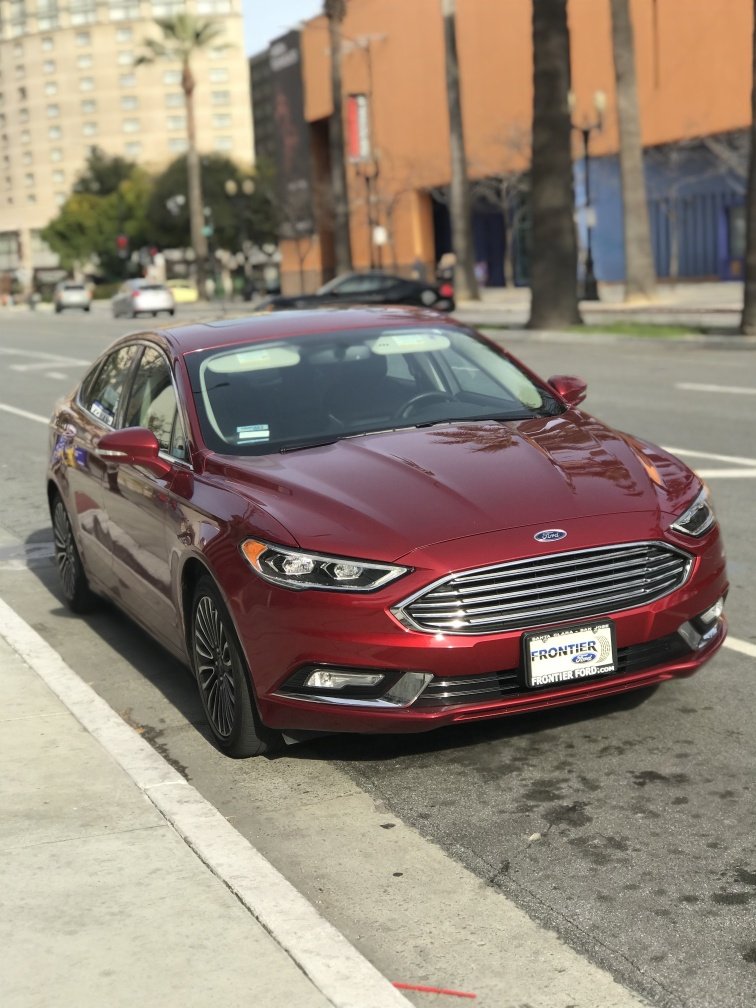} &
		\includegraphics[width=0.3\linewidth,clip,trim=0 125 0 125]{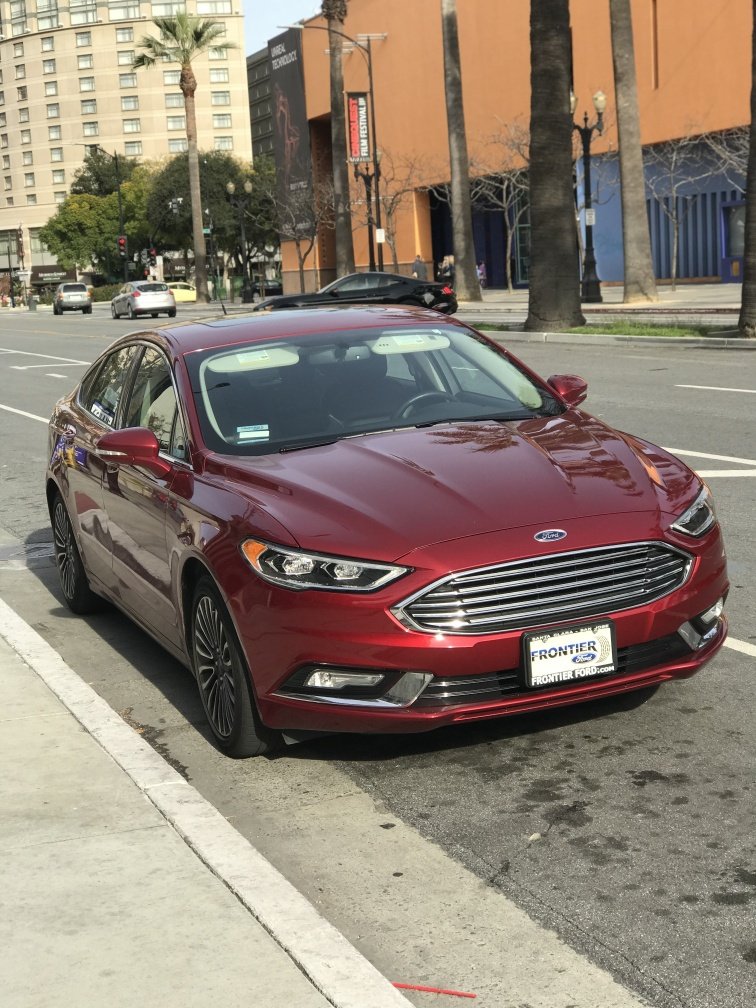}\\
		\\
		Input & Ours & iPhone Portrait
	\end{tabular}
	\caption{Qualitative comparison of our method and iPhone portrait mode (Cropped). In the first row, our method achieves comparable results against iPhone. In the second row, iPhone fails to render shallow DoF effects.}
	\label{fig:qualiphone}
\end{figure}

\subsection{User Study}
\label{sec:userstudy}
To further evaluate the perceptual quality of our results, We conducted two user studies.
In the first user study, we compare our results to the photos captured with iPhone portrait mode. 
Since we cannot control the focal depth in iPhone photos, and we cannot create defocus foreground images, we generate our results by manually matching focus parameters. 
In the second user study, we compare ours against MegaDepth\cite{li2018megadepth} + \cite{yang2016virtual} (\emph{i.e.} the state-of-the-art depth prediction method + RGB-D lens blur rendering) on images downloaded from the internet, which are captured by different cameras with various objects and scenes. 

\begin{figure}[t]
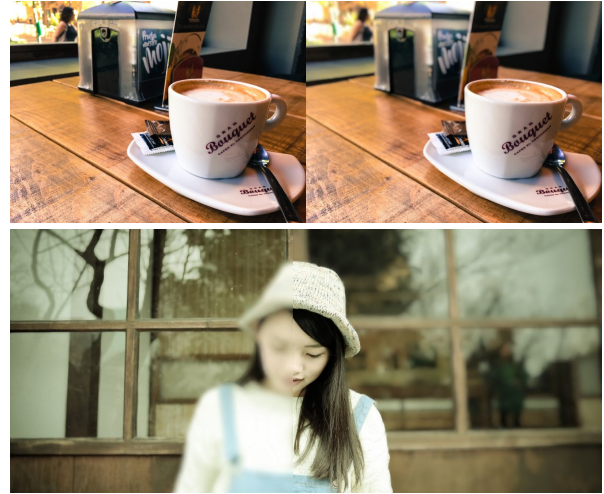

    \centering
    \tabcolsep0.3pt \renewcommand{\arraystretch}{0.5}
    \begin{tabular}{cc}
    \includegraphics[height=2.94cm,clip,trim=0 0 0 0 ]{{{figures/aperture/29117126306_d1c4013fd7_k}}} &
    \includegraphics[height=2.94cm,clip,trim=0 0 0 0]{{{figures/aperture/29117126306_d1c4013fd7_k_1}}} \\
    \end{tabular}
    \begin{tabular}{cc}
    \includegraphics[height=3.5cm,clip,trim=0 125 534 100]{{{figures/qual_results/12191004846_8c4ae50774_k_kr_0.812_df_0.155_x_672_y_462}}} &
    \includegraphics[height=3.5cm,clip,trim=489 125 0 100]{{{figures/qual_results/12191004846_8c4ae50774_k_kr_0.812_df_0.904_x_982_y_594}}} \\
    \end{tabular}
    \caption{\label{fig:refocusing}Refocusing allows us to change both the synthetic aperture size (top) and the focal plane (bottom). \textit{Photo credits: DAVID BURILLO and Greg Tsai.}}
\end{figure}

These two studies consisted of 86 and 58 participants respectively, During the study, each participant was shown 20 image sets, each consisting of the input all-in-focus image and the two shallow DoF images. 
All images and results in the study are included in the supplementary materials.
In each image set, participants were asked which result they preferred, or if there was no preference. 
Fig.~\ref{fig:userstudy} shows the results of this experiment, we can see that when compared to the iPhone portrait mode, our method performs comparably, despite the fact that our method uses \emph{predicted} depth while iPhone uses measured depth maps.
In the second user study, our approach significantly outperforms the MegaDepth~\shortcite{li2018megadepth} + \cite{yang2016virtual}.
We note that the iPhone portrait mode only works in a very limited depth range, which has been heavily optimized for (Fig.~\ref{fig:qualiphone} shows two examples), whereas in the second user study, the scenes are more diverse, and both competitors only rely on a single image as input. 
Fig.~\ref{fig:refocusing} shows results of changing the focus parameters on the same image. 

\begin{figure}
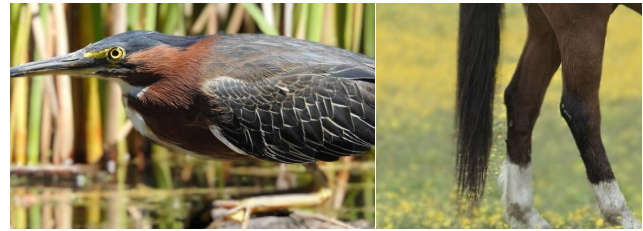

    \centering
    \tabcolsep0.3pt \renewcommand{\arraystretch}{0.5}
    \begin{tabular}{cc}
    \includegraphics[height=3.0cm,clip,trim=350 150 00 200]{{{figures/failure/19972543462_9144f93908_k_kr_0.495_df_0.874_x_1156_y_612}}} &
    \includegraphics[height=3.0cm,clip,trim=100 150 600 300]{{{figures/failure/26847694287_eec6a2b0b6_k_kr_0.754_df_0.902_x_1068_y_356}}} \\
    \end{tabular}
    \caption{\label{fig:limitations}Failure cases. The leg of the bird, and part of the grass are incorrectly blurred and focused. \textit{Photo credits: Eric Sonstroem and C Watts.}}
\end{figure}

\subsection{Limitations and Future Work}
Our method has several limitations. 
First, single image depth estimation is still a challenging problem, and errors in the depth computation are reflected in the rendered DoF results (Fig.~\ref{fig:limitations}).
Therefore, one of the future works will be improving our depth prediction module by using larger training datasets and better model architectures.

Second, we note that most computation of our method has to be performed at a low resolution due to current GPU memory limitations.
To address this problem, we proposed a recurrent guided upsampling module which allows us to generate arbitrarily high resolution outputs at test time. However, it may further improve the performance if we can optimize the whole system directly on high resolution DoF images. 

Finally, we note that it leads to a decrease in quality if we jointly train all the modules on our current datasets. 
\final{The reasons are twofold: i) Rendering shallow DoF images with existing methods based on the collected depth leads to unpleasant artifacts, so any network trained with the rendered results as ground truth simply learns to recreate the artifacts in these existing approaches; ii) The synthetic dataset is artifact-free. However, training on this dataset in an end-to-end manner results in poor generalization ability to real scenes.}
We believe that a large representative dataset with high quality real-world shallow DoF images could be used to jointly train all the modules in our system for improved results, if such a dataset could be captured on a similar variety of dynamic scenes.

\section{Conclusion}
We have presented a neural network architecture,  which combined with carefully crafted datasets consisting of both real and synthetic images, is capable of generating high quality shallow Depth-of-field results from a single RGB image. With the introduction of the lens blur module in low-res feature space and the guided upsampling module that can be recurrently applied to increase resolution, our model can generate high-resolution shallow DoF images with controllable focal depth and aperture size at interactive rates.


\begin{acks}
This work is supported by Adobe Research and the National Natural Science Foundation of China under Grant 61725202, 91538201 and 61751212.
We thank Flickr users mentioned in the paper whose photos are under Creative Commons License.
\end{acks}

\bibliographystyle{ACM-Reference-Format}
\bibliography{src/shallowdof}

\end{document}